\documentclass{article} %
\usepackage{iclr2027_conference,times}

\usepackage{hyperref}
\usepackage{url}

\usepackage[utf8]{inputenc} %
\usepackage[T1]{fontenc}    %
\usepackage{booktabs}       %
\usepackage{amsfonts}       %
\usepackage{nicefrac}       %
\usepackage{microtype}      %
\usepackage{xcolor}         %
\usepackage{bbm}
\usepackage{float}

\usepackage{multirow}
\usepackage{algorithm}
\usepackage{algpseudocode}
\usepackage{enumitem}
\usepackage{amsmath, amssymb}
\usepackage{mathtools}
\usepackage{graphicx}
\usepackage{comment}

\usepackage{amsthm}
\newcommand{\eat}[1]{}
\usepackage{siunitx}
\usepackage{subcaption} 

\newtheorem{proposition}{Proposition}

\title{Hearing the Unspoken: Language Model Priors for Acoustic Adversarial Attacks}

\author{%
   Jiani Xie \\
   University of Melbourne \\
   \texttt{jiani.xie1@unimelb.edu.au} \\
   \And
   Andrew C. Cullen \\
   University of Melbourne \\
   \And
    \hspace{2.95cm}
    \AND
   Paul Montague \\
   DST Group \\
   \And
   Benjamin I. P. Rubinstein \\
   University of Melbourne \\
}

\iclrfinalcopy %
\begin{document}

\maketitle

\begin{abstract}
Automatic Speech Recognition (ASR) systems operating in real-time settings must process acoustic input under strict temporal constraints, where transcription decisions are inherently made on incomplete information. This causal constraint serves as an information bottleneck on attackers, significantly limiting attack performance. Our new Semantic Gambit attack overcomes this causal limitation by augmenting the adversary with predictive context derived from a Large Language Model in real-time. Our experiments show that this form of augmentation can elevate the corpus-level Word Error Rate to 35.6\%---a three-fold increase over the current state-of-the-art streaming attack. Ultimately, this work reveals how common, low-latency LLM tooling can be exploited to systematically subvert real-time ASR pipelines.
\end{abstract}

\section{Introduction}
Modern Automatic Speech Recognition (ASR) systems are vulnerable to adversarial perturbations, which, when overlaid on speech, degrade the resulting transcription~\citep{carlini2018audio_adversarial_examples, olivier2023transferable, cheng2024alif}. Intuitively, attack performance is proportional to the information available to the attacker---the more the attacker knows about the target utterance, the more precisely the perturbation can be shaped to disrupt it. In offline settings, this is no obstacle, since the attacker can leverage the full utterance.

However, many attack scenarios require constructing attacks in situ, where the attacker can only access information from \emph{before} the attack is deployed~\citep{chiquier2022voice_camouflage}. The moment the perturbation is injected, it distorts the acoustic environment, foreclosing further clean observation and bounding accessible information.

\begin{figure}[t]
  \centering
  \includegraphics[width=0.7\linewidth]{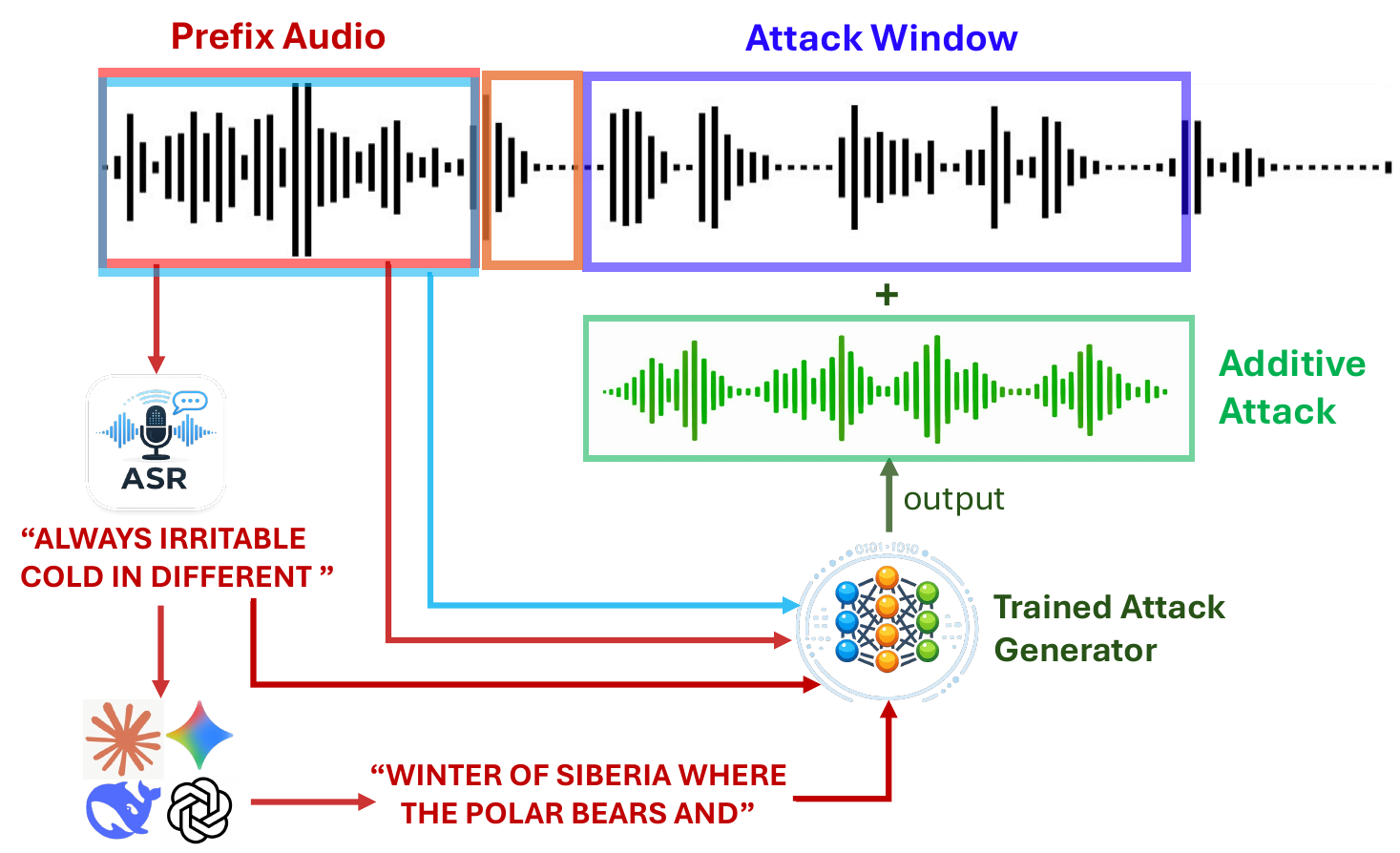}
  \caption{\textbf{Overview of Attack Pathways}. In streaming attacks, a prefix audio segment is exploited by a generator to construct an attack in the attack window. Our SG attack enhances information availability by supplementing audio information with prefix audio, an ASR transcript of the prefix, and an LLM forecast (red arrows), producing an information advantage over the current SOTA for such attacks (blue arrow). The orange box denotes the computational delay for synthesizing attacks.}
  \label{fig:teaser}
\end{figure}

While future speech cannot be heard, the partial transcript of what has already been spoken is often sufficient for a Large Language Model (LLM) to predict what comes next. We exploit this observation to circumvent the information bound with our \textbf{Semantic Gambit} (SG) attack. A multimodal generator then fuses this predicted continuation with the acoustic prefix to produce an input-specific perturbation in real time (Figure~\ref{fig:teaser}). To the best of our knowledge, this is the first attack framework to use LLM-based predictions to guide perturbation generation in a different modality.

At its strongest configuration, the system elevates the victim model's word error rate from 2\% to over 35\% under a realistic 20 dB SNR constraint, approximately three times the degradation achieved by the state-of-the-art streaming attack~\citep{chiquier2022voice_camouflage}. %
Ablations confirm that semantic conditioning contributes up to 20 percentage points beyond what acoustic context alone provides, and the full pipeline generates a 3-second perturbation in just 0.6 seconds on average, operating at roughly 5× real-time without specialist optimization---fast enough to attack live speech. %
These results demonstrate that the community may be significantly underestimating ASR risks, and that AI risks should consider the potential of augmented information.

\section{Related Work}
\label{sec:related_work}

\paragraph{Audio-Specific Challenges}
Audio adversarial attacks differ from image attacks in ways that complicate both design and evaluation. Standard $\ell_p$ distances correlate poorly with perceived audibility, motivating perceptually grounded budgets such as 
psychoacoustic masking constraints \citep{qin2019imperceptible_asr, schonherr2019psychoacoustic_hiding, szurley2019perceptual_audio_attacks}. ASR systems also operate on diverse input representations, from raw waveforms to Mel spectrograms~\citep{jung2019rawnet, chan2016_listen_attend_spell}, and a perturbation that appears small in one can become salient in another~\citep{hussain2021waveguard}. Deployed pipelines also add microphone, room, and codec effects with no analogue in image settings~\citep{yakura2018robust_audio_physical}, and streaming deployments impose causal constraints. %

\paragraph{Attacks on ASR Systems}
Early attacks on ASR pursued offline, targeted perturbations against end-to-end models~\citep{carlini2018audio_adversarial_examples, yakura2018robust_audio_physical, yuan2018commandersong}, followed by universal variants that generalize across utterances~\citep{neekhara2019universal_audio, abdoli2019universal_adversarial_audio} and later incorporated EoT-based robustness and psychoacoustic objectives~\citep{zong2021targeted_uap_isc, sun2024commanderuap}. %
Recent studies on Wav2Vec 2.0 and Whisper typically adopt standard gradient-based objectives within larger pipelines~\citep{olivier2023transferable}, 
and some sidestep transcription attacks entirely by manipulating task-level behavior, such as forcing Whisper to translate rather than transcribe~\citep{raina2024controlling_whisper}. 
Few of these methods employ generative networks that map inputs to utterance-specific perturbations in a single forward pass. Under causal streaming constraints, \citet{gong2019realtime} train a generator to emit perturbations on streaming input, though for keyword classification rather than speech recognition; the closest such work on ASR, Neural Voice Camouflage~\citep{chiquier2022voice_camouflage}, conditions solely upon acoustic information.

\paragraph{Modern ASR Architectures}
The ASR landscape has shifted rapidly in ways relevant to adversarial robustness. DeepSpeech 2~\citep{amodei2016deepspeech2} paired recurrent encoders with a CTC~\citep{graves2006ctc} objective; Wav2Vec 2.0~\citep{baevski2020wav2vec2} introduced self-supervised Transformer-based pre-training and became the de facto acoustic backbone; Whisper~\citep{radford2022whisper} scaled to encoder-decoder generation trained on roughly 680k hours of weakly supervised multilingual audio. Each generation brings different inductive biases and decoding mechanisms, so vulnerabilities established on one do not necessarily transfer to another. The present work targets Wav2Vec 2.0, whose clean WER on LibriSpeech test-clean is below 2\%.

\paragraph{Language Models as Auxiliary Components in Adversarial Systems}
In prior works, LLM-generated paraphrases and synonym substitutions have been deployed against NLP classifiers~\citep{li2020bertattack}, and stronger LLMs have served as autonomous red-team actors, that iteratively refine jailbreak prompts against other models~\citep{chao2023jailbreaking}. LLMs have also been adopted as evaluators during training and assessment~\citep{zheng2023judging}, though this practice carries known issues with inconsistency, positional bias, and prompt sensitivity~\citep{wang2023large_models_fair_evaluators}. In each of these roles, however, the language model either operates within the same modality as the attack or serves as an evaluation proxy. To our knowledge, no prior work has used a language model's predictive capabilities  %
to guide cross-modal attacks. %

\section{An Acoustic Attack Taxonomy}
\label{sec:problem_setting}

An unacknowledged but key taxonomical distinction between attacks on ASR systems relates to the attacker's knowledge, where increasing both the available information and the relevance of the information yields stronger attacks. Based upon this observation, we begin by classifying attacks in terms of an \emph{information filtration operator} that determines what portion of the target utterance is visible to the generator.

\paragraph{General Formulation.}
Let $\mathcal{X} \subseteq \mathbb{R}^T$ denote the space of valid acoustic waveforms %
and let $\mathcal{Y}$ be the space of token sequences. An ASR model is defined as the mapping $M : \mathcal{X} \to \mathcal{Y}$, which assigns a transcription $y = M(x)$ to an input $x \in \mathcal{X}$. %
An adversarial attack seeks a perturbation $\delta \in \mathcal{X}$ to maximize the divergence between the original transcription and the perturbed result $y' = M(x + \delta)$, typically measured via Word Error Rate (WER). To formalize the bounds on the attacker's knowledge, we rely on the concept of a filtration---which represents the total information accumulated by the process up to a specific moment. Let $\mathcal{F}_{t} = \sigma(x_1, \ldots, x_t)$ denote the filtration of the clean signal and $\mathcal{F}_{t}^{W} = \sigma(w_1, \ldots, w_t)$ the observed filtration of the perturbed signal $w = x + \delta$ at time $t$.

We characterize the attacker's capability through an information operator $H : \mathcal{X} \to \mathcal{Z}$, where $\mathcal{Z}$ denotes the observation space. A learned attack mechanism $f_\theta : \mathcal{Z} \to \mathcal{X}$ produces the perturbation $\delta = f_\theta(H(x))$. The adversarial objective is thus:
\begin{equation}
    \max_{f_\theta \in F(H)} \;\; \text{WER}\bigl(M(x), M(x + f_\theta(H(x)))\bigr)\enspace, 
\end{equation}
where $F(H)$ represents the class of appropriate structural constraints.%

\paragraph{Perfect Information.}
The most permissive regime grants the attacker full access to the utterance, $H(x) = x$. This characterizes Projected Gradient Descent (PGD)-style attacks~\citep{carlini2018audio_adversarial_examples} and offline methods that optimize over the entire signal~\citep{qin2019imperceptible_asr, cheng2024alif}. Here, the generator can iteratively refine $\delta$ against the victim model. However, these attacks are \emph{retroactive} by construction: by the time optimization ends, the utterance has already been spoken.

\paragraph{Hidden Information.}
At the opposite extreme, universal attacks discard input dependence entirely, $H(x) = \varnothing$. A single perturbation is precomputed offline and applied to every utterance~\citep{neekhara2019universal_audio, sun2024commanderuap}. While sidestepping temporal constraints, this forces the perturbation to optimize for average-case degradation across the training distribution, foreclosing access to input-dependent victim vulnerabilities. %

\paragraph{Partial Information.}
Real-world threats %
exist in the space between these extremes, where the attacker's knowledge is limited by causal and environmental factors. In streaming settings, for example, the presence of the perturbation $\delta$ induces a form of deleterious self-interference. As it is assumed that the attacker's perturbation induces a non-invertible physical channel transformation, the underlying clean signal is obscured in an unrecoverable manner. This forces the information histories to diverge such that $\mathcal{F}_{t}^{W} \subset \mathcal{F}_{t}$, whereby $\delta \neq 0$ must induce a degradation in conditional entropies  $\mathcal{H}(x_t | \mathcal{F}_{t}^{W}) > \mathcal{H}(x_t | \mathcal{F}_{t})$. To practically model this, we define a \emph{causal barrier} $t_p$ where $\delta_{t \leq t_p} = 0$.

\begin{proposition}[Information Admissibility]
\label{prop:admissibility}
An attack is admissible with respect to a causal barrier $t_p$ if the operator $H(x)$ is $\mathcal{F}_{t_p}$-measurable. Crucially, while $t_p$ bounds what the attacker can \textbf{measure}, it does not bound the set of \textbf{computable functions} of those measurements. For any $\mathcal{F}_{t_p}$-measurable function $\phi$, the operator $H(x) = [x_p, \phi(x_p)]$ is admissible.
\end{proposition}

To formalize this, we assume that the attacker observes the prefix $H(x; t_p) = x \cdot \mathbbm{1}[t < t_p]$. Following the Neural Voice Camouflage attack of \citet{chiquier2022voice_camouflage}, we also assume that the attack requires a computation delay $\tau$, and targets a window of duration $\Delta$. The resulting perturbation is thus
\begin{equation}
    \delta_t = f_{\theta,t}\bigl(H(x; t_p)\bigr), 
    \quad t \in \{t_p + \tau, \ldots, t_p + \tau + \Delta\}\enspace.
\end{equation}
This framework captures the fundamental tension of real-time ASR attacks: the necessity of acting upon partial history while managing the informational loss induced by the attack itself.

\paragraph{Attack Constraints.}
In addition to the informational restriction imposed by $H$, the perturbation is typically constrained to be imperceptible to a human listener. In the vision literature, this is conventionally expressed as a magnitude bound on $\delta$ under an $\ell_p$ norm, $\|\delta\|_p \le \epsilon$. In the audio domain, however, $\ell_p$ bounds correlate poorly with human audibility~\citep{qin2019imperceptible_asr, schonherr2019psychoacoustic_hiding}: the human ear is sensitive to energy rather than per-sample magnitude, so a perturbation that is small under $\ell_p$ may still be clearly audible while one that is large under $\ell_p$ may be masked entirely by the signal. 
We therefore adopt the signal-to-noise ratio,
\begin{equation}
\mathrm{SNR}(\delta, x) = 10 \log_{10}\!\bigl(\|x\|_2^2 \,/\, \|\delta\|_2^2\bigr) \;\ge\; \mathrm{SNR}_{\text{tgt}}\enspace,
\end{equation}
which constrains perturbation energy relative to the energy of the signal segment being perturbed. The same constraint applies to every method evaluated.

\section{Semantic Gambit: An Augmented Information Attack}
\label{sec:augmented_information}

Our information-driven taxonomy begets the question: if attack performance is correlated with information, can an attacker improve the volume of information available to them without violating causal barriers? %
This is not as impossible a task as it sounds. While the acoustic channel is bound by the self-interference that increases entropy $\mathcal{H}(x_t | \mathcal{F}_{t}^{W})$, LLMs, by their nature, are trained to construct viable snapshots of the future. As such, we extend the partial-information operator to expose both the acoustic prefix and a predicted linguistic continuation:
\begin{equation}
H(x) \;=\; \bigl[\, x_p,\; \tilde{T}(x)\,\bigr], 
\qquad 
\tilde{T}(x) \;=\; M(x_p) \,\oplus\, L\bigl(M(x_p)\bigr)\enspace,
\end{equation}
where $x_p$ is the observed prefix, $L$ is an LLM for future-text prediction, and $\oplus$ denotes concatenation. 

\begin{proposition}[Symbolic Invariance]
\label{prop:invariance}
Let $\delta$ be a perturbation emitted at time $t > t_p$. The textual channel $\tilde{T}(x)$ is invariant to $\delta$, whereas the acoustic filtration $\mathcal{F}_t^W$ is a function of $\delta$.
\end{proposition}

The value of this invariance is central to our method. The acoustic channel is bound by the temporal requirement that the attacker cannot hear what has not been spoken, and cannot reliably discern information after the perturbation has been emitted. The textual channel $\tilde{T}(x)$, however, is purely $\mathcal{F}_{t_p}$-measurable; it requires no further audio and is immune to the acoustic degradation induced by the emitted perturbation. Prior streaming attacks~\citep{gong2019realtime, chiquier2022voice_camouflage} have implicitly restricted $\phi$ to the identity, equating ``available information'' with ``observed audio.'' Semantic Gambit operates beyond these bounds by using pre-trained linguistic priors that an acoustic-only generator cannot access. \textbf{It breaks the information barrier not by hearing more, but by knowing more.}

\subsection{Implementation}
\label{sec:implementation}

We now describe the concrete realization of the augmented operator $H$ introduced in Section~\ref{sec:augmented_information}. %
Each utterance $x$ is partitioned into four contiguous segments: the \emph{prefix} signal $x_p$ of length $t_p$, during which the attacker passively observes; the \emph{delay} of length $\tau$, modeling the computational latency of the real-time pipeline; the injection-window signal $x_\star$ of fixed length $\Delta_\star = 3\,\text{s}$, into which the perturbation is injected; and the suffix which corresponds to the remaining audio. The perturbation $\delta$ is then defined over the same temporal window as $x_{\star}$. The augmented filtration operator $H(x) = [x_p,\, \tilde{y}_\star]$, where $\tilde{y}_\star$ is the ASR-transcribed prefix concatenated with an LLM-predicted continuation, depends on $x$ only through the prefix $x_p$.%

\begin{algorithm}[t]
\caption{\textbf{Semantic Gambit (SG) Training Procedure.}}
\label{alg:semantic_gambit}
\begin{algorithmic}[1]
\Require Victim ASR $M$, language model $L$, generator $G_\theta$, training set $\mathcal{D}$, target SNR $\mathrm{SNR}_{\text{tgt}}$
\Ensure Optimized generator parameters $\hat\theta$
\For{$\text{epoch} = 1, \dots, E$}
    \For{each utterance $(x, y^{\text{gt}}) \in \mathcal{D}$}
        \State Partition $x$ into $x_p$ and $x_\star$ per the fixed temporal schedule
        \State $y_p \leftarrow M(x_p)$;\quad $y_f \leftarrow L(y_p)$ \Comment{Semantic expansion}
        \State $\tilde\delta \leftarrow G_\theta\bigl(\text{MFCC}(x_p),\; E_{\text{sem}}(y_p \oplus y_f)\bigr)$ \Comment{Perturbation generation}
        \State $\delta \leftarrow \textsc{ScaleToSNR}(\tilde\delta,\, x_\star,\, \mathrm{SNR}_{\text{tgt}})$ \textsuperscript{$\dagger$} \Comment{Constraint enforcement}
        \State $\theta \leftarrow \theta - \eta\,\nabla_\theta\bigl(-\mathcal{L}_{\text{CTC}}(M(x + \delta),\, y^{\text{gt}})\bigr)$ \Comment{Optimization}
    \EndFor
\EndFor
\State \Return $\hat\theta$
\end{algorithmic}
{\footnotesize $\dagger$\,The \textsc{ScaleToSNR} operator applies
a uniform scalar rescaling to match a prescribed perturbation
budget; it does not feed target-window information into the
generator. See Appendix~\ref{app:training_details},
\S\,\emph{Perturbation scaling}.}
\end{algorithm}

\paragraph{Attack Pipeline.}
The generator is a composition of three maps acting on the prefix. The victim ASR $M$ produces a partial transcript $y_p$; the language model $L$ extends it into a forecast $y_f$ of upcoming linguistic content, so that $y_p \oplus y_f$ provides the text channel; and the multimodal generator $G_\theta$ fuses acoustic features extracted by $\text{MFCC}(\cdot)$ from the prefix audio with semantic embeddings produced by $E_{\text{sem}}(\cdot)$ from the concatenated token sequence, yielding the raw perturbation $\tilde\delta$ supported on the target interval, i.e., the time window into which the perturbation is injected. $\delta$ is then produced by rescaling to the target SNR (relative to $x_\star$), with end-to-end training maximizing Connectionist Temporal Classification (CTC) loss~\citep{graves2006ctc} against the ground-truth transcript $y^{\text{gt}}$. %

\paragraph{Generator Architecture.}
$G_\theta$ has two stages. Stage 1 applies multi-modal self-attention jointly over audio frames and text tokens, letting the network learn which aspects of the linguistic forecast most inform the perturbation. Stage 2 is a Perceiver~\citep{jaegle2021perceiver} that compresses this fused representation into a fixed-dimensional latent set, from which the perturbation waveform is decoded. Joint attention over both channels lets the perturbation be shaped by past ($y_p$) and predicted ($y_f$) speech simultaneously. Appendix~\ref{app:transformer_architecture} compares this design with three alternative fusion strategies.

\section{Experiments}
\label{sec:experiments}
We evaluate Semantic Gambit along two axes. First, \emph{attack efficacy across information regimes}: we benchmark SG against representatives of each regime from Section~\ref{sec:problem_setting}---a hidden-information universal perturbation and a white-noise control at the floor, an offline PGD attack with perfect-information access at the ceiling, and partial-information methods that share the streaming constraint but differ in the information operator $H$. This design tests whether linguistic forecasting closes the gap to perfect-information performance without violating causality, decomposes any gain into acoustic, architectural, and semantic components, and verifies real-time feasibility via end-to-end latency. Second, \emph{dataset transferability}: whether a generator trained on one corpus retains efficacy on held-out distributions. This isolates generalizable adversarial structure from corpus-specific overfitting and bounds the attack's practical reach beyond the training distribution.

\eat{We evaluate Semantic Gambit along two axes. First, \emph{attack efficacy across information regimes}: we benchmark SG against representatives of each regime defined in Section~\ref{sec:problem_setting} -- a hidden-information universal perturbation and a white-noise control at the floor, an offline PGD attack with perfect-information access at the ceiling, and a family of partial-information methods that share the streaming constraint while differing in the content of information operator $H$. This design tests whether augmenting the partial-information regime with linguistic forecasting closes the gap toward perfect-information performance without violating causality, and decomposes any gain into its acoustic, architectural, and semantic components. Our second testing axis focuses upon \emph{dataset transferability}: whether a generator trained on one corpus retains its efficacy when deployed against held-out distributions. We also report pipeline latency to confirm that the full system operates within the real-time budget imposed by the streaming setting.}

\subsection{Experimental Setup} \label{subsec:setup}

Our primary experiments train and evaluate on the LibriSpeech corpus~\citep[CC~BY~4.0 License]{panayotov2015librispeech}: train-clean-100 (approximately 28{,}500 utterances) for training, dev-clean for validation and model selection, and test-clean for evaluation. Cross-dataset experiments additionally use Common Voice 25.0 English~\citep[CC0~1.0 License]{ardila2020common_voice}, training a separate generator under matched compute (Section~\ref{sec:cross_transfer}). The default victim is a Wav2Vec 2.0 Large model~\citep[MIT License]{baevski2020wav2vec2}, pre-trained on Libri-Light (LV-60K) and fine-tuned with CTC on 960\,h of LibriSpeech, achieving approximately 2\% clean WER on test-clean; cross-model experiments substitute
HuBERT-Large~\citep[Apache~2.0 License]{hsu2021hubert} and Whisper-small~\citep[Apache~2.0 License]{radford2022whisper} at evaluation time without
retraining the generator (Section~\ref{sec:cross_model}). The generator $G_\theta$ contains approximately 7.3M parameters. Semantic forecasts are provided by Llama 3 8B~\citep[Meta Llama~3 Community License]{dubey2024llama3} quantized to 8-bit precision, serving as the language model $L$ that produces up to 15 continuation tokens per prefix transcript $y_p$, which are decoded and re-tokenized at the character level before entering $G_\theta$ (see Appendix~\ref{app:training_details}). All attacks are evaluated under a 20\,dB SNR constraint. 

We systematically vary two temporal parameters: the prefix length $t_p \in \{1.0,\, 2.0,\, 3.0,\, 4.0\}\,\text{s}$, which controls the volume of acoustic and linguistic context available to the attacker before it acts, and the injection delay $\tau \in \{0.0,\, 0.5,\, 1.0\}\,\text{s}$, which controls the recency of that information relative to the point at which the perturbation is injected. Concretely, $t_p$ governs how much context the attacker can extract from the prefix, while $\tau$ represents the processing time required to access the acoustic signal and pass it through the pipeline before the perturbation is ready. The range of tested delays allows us to understand the influence of information recency on attack performance; Appendix~\ref{app:latency} reports hardware measurements confirming that the actual pipeline latency falls within this range. Together, this grid tests whether the value of linguistic forecasting persists as operational conditions tighten.

\subsection{Baselines}
\label{subsec:baselines}
We compare Semantic Gambit against five reference points spanning the information regimes of Section~\ref{sec:problem_setting}. \textbf{White Noise}, matched in duration and SNR to our perturbations, verifies that degradation stems from adversarial optimization rather than simple acoustic masking. 
\textbf{AO*} reimplements the purely acoustic convolutional generator of~\citet{chiquier2022voice_camouflage}, the closest prior attack on ASR in the streaming input-specific setting and therefore our state-of-the-art reference point.
\textbf{AO} retains SG's full generator architecture and training procedure but zeroes the text channel, suppressing both the prefix transcription and the LLM continuation; any gap to SG is therefore attributable to linguistic anticipation rather than architectural differences. 
\textbf{Universal} applies a single input-agnostic perturbation to every utterance, establishing a lower bound in the absence of input-specific information. Prior universal audio perturbations target a different ASR victim~\citep{neekhara2019universal_audio} or audio classifiers~\citep{abdoli2019universal_adversarial_audio}, so we optimize one directly against our victim under our SNR budget.
\textbf{GT} replaces the LLM-generated forecast with the victim ASR's own transcript of the target window, an oracle for the forecast SG's LLM attempts to produce.
\textbf{PGD}~\citep{madry2018pgd_adversarial_training} optimizes the perturbation with non-causal \emph{access to the complete utterance}, providing a ceiling for purely gradient-based attacks that violate the streaming constraint. Implementation details and hyperparameters are deferred to Appendix~\ref{app:baseline_details}.

\subsection{Main Results}
\label{subsec:results}
\paragraph{Overall Efficacy.}
Semantic Gambit achieves a roughly 17-fold increase in corpus-level WER over clean (2.05\% $\to$ 35.63\%) at its strongest operating point under a 20\,dB SNR constraint, with sentence-level mean WER reaching 49.6\% (Table~\ref{tab:sentence_ci}). At every zero-delay configuration, SG exceeds even the offline PGD
baseline despite PGD having non-causal access to the complete waveform and ground-truth transcript during optimization. SG operates under a strictly causal constraint---only the prefix audio is available at generation time---yet surpasses PGD, demonstrating that the linguistic channel compensates for the information deficit imposed by the streaming setting. This performance difference is indicative of the utility of providing the attacker with LLM-generated context, as this additional information is enough to surpass the performance of an attack (PGD) that is able to look past the causal barrier. Furthermore, the white-noise control confirms that these gains reflect learned adversarial structure rather than acoustic masking.

\begin{table*}[htbp]
\centering
\small
\caption{Corpus-level WER (\%) under a 20\,dB SNR constraint, with a 3.0\,s target window. White noise produced negligible degradation (within 0.2\% of clean) and is omitted. Bold marks the strongest attack per row among streaming-comparable methods (SG, AO, AO*, Universal). Note that GT and PGD break the causality barrier, and as such are not directly comparable to the prefix attacks.}
\label{tab:baselines_corpus_20db}
\begin{tabular}{@{}cc c c cc cc c@{}}
\toprule
\textbf{Prefix (s)} & \textbf{Delay (s)} & \textbf{Clean} & \textbf{SG} & \textbf{AO} & \textbf{AO*} & \textbf{Universal} & \textbf{GT (Oracle)} & \textbf{PGD} \\ \midrule
\multirow{3}{*}{1.0} 
 & 0.0 & 2.05 & \textbf{35.63} & 15.15 & 12.23 & 17.37 & 23.53 & 17.01 \\
 & 0.5 & 2.05 & 16.26 & \textbf{19.57} & 8.28 & 14.91 & 16.00 & 15.64 \\
 & 1.0 & 2.02 & \textbf{18.89} & 14.52 & 5.34 & 17.21 & 21.42 & 14.58 \\
 \midrule
\multirow{3}{*}{2.0} 
 & 0.0 & 2.02 & \textbf{32.72} & 12.58 & 2.16 & 14.45 & 19.01 & 14.46 \\
 & 0.5 & 2.01 & \textbf{20.32} & 14.17 & 4.38 & 10.46 & 14.51 & 13.75 \\
 & 1.0 & 2.03 & \textbf{18.95} & 13.42 & 3.19 & 10.62 & 23.40 & 13.17 \\
 \midrule
\multirow{3}{*}{3.0} 
 & 0.0 & 2.03 & \textbf{22.69} & 14.82 & 4.27 & 13.14 & 16.47 & 13.21 \\
 & 0.5 & 2.02 & \textbf{21.17} & 19.46 & 2.35 & 18.52 & 20.00 & 12.70 \\
 & 1.0 & 2.01 & \textbf{11.32} & 11.11 & 8.45 & 9.99 & 12.94 & 12.02 \\
 \midrule
\multirow{3}{*}{4.0} 
 & 0.0 & 2.01 & \textbf{20.57} & 19.85 & 5.80 & 10.05 & 12.44 & 12.18 \\
 & 0.5 & 1.98 & \textbf{19.12} & 15.03 & 4.84 & 11.89 & 14.27 & 12.13 \\
 & 1.0 & 1.97 & \textbf{18.68} & 17.68 & 8.78 & 11.15 & 14.46 & 11.56 \\ \bottomrule
\end{tabular}%
\end{table*}

\paragraph{Information Hierarchy and Attack Effectiveness.}
The three configurations, Universal, Acoustic-Only (AO), and Semantic Gambit, occupy progressively richer positions along the information axis constraining real-time ASR attacks, and their performance traces this hierarchy directly. At zero delay, averaging across prefix lengths, transitioning from the utterance-agnostic Universal attack (13.8\% WER) to AO (15.6\%) induces a modest gain, while transitioning from AO to SG (27.9\%) yields a 79\% relative increase, confirming that linguistic augmentation is the decisive factor. Universal's ceiling is structural: a fixed perturbation forecloses access to input-dependent vulnerabilities, and the per-utterance signals it sees during training are averaged across the
distribution, diluted past the point where any single input meaningfully shapes the perturbation. AO recovers per-utterance adaptivity, but acoustics are only one layer of what audio carries: AO observes the waveform alone, blind to the lexical and semantic structure that lives in the same signal. SG's linguistic priors widen the information channel along two axes: the LLM extrapolates likely future content, and the prefix transcription itself supplies the generator with explicit linguistic context for crafting the perturbation. The augmentation pays off most at the strongest operating point (1.0\,s prefix, zero delay), where SG yields a 135\% relative increase over AO (35.63\% vs.\ 15.15\%), confirming that what governs real-time attack effectiveness is not what the attacker hears, but what it already knows.

\paragraph{Effect of Prefix Length.}
Across methods, attack efficacy decreases as the observed prefix grows longer: SG drops from 35.63\% WER at a 1.0\,s prefix to 20.57\% at 4.0\,s under zero delay, and the same trend appears in GT, PGD, and AO*. Since PGD uses no text conditioning and AO* uses no generator network, the effect cannot be attributed to the text pipeline alone. Two mechanisms appear to compound. First, longer prefixes give the victim a cleaner context to anchor its decoding, making later frames harder to perturb regardless of attack origin; this accounts for the baseline slope visible in PGD and AO*. Second, for text-conditioned attacks, the forecast's value depends on target predictability, which degrades with position as accumulated branching uncertainty outpaces the benefit of additional prefix conditioning. 
AO faces the victim-side mechanism but bypasses the text channel entirely, and does not exhibit the same decline across prefix lengths, consistent with a text-specific component. Quantitatively, SG's advantage over AO narrows from 20.5 points at 1.0\,s to under 1 point at 4.0\,s.
Comparisons between methods hold within each row, where all methods share the same evaluation subset; across rows, the duration filter means longer prefixes are evaluated on progressively longer utterances, so the trend combines prefix length with utterance length (Appendix~\ref{app:cross_matched_subset}).
We examine architectural and training considerations in Appendix~\ref{sec:appendix_prefix}.

\paragraph{Sensitivity to Injection Delay.} Increasing delay forces the perturbation to target audio further removed from the last observed context without providing any compensating information, producing commensurate decreases in performance. We stress that the purpose of the delay sweep is not to seek a better operating point but to characterize the framework's robustness under realistic computational latency, since any deployed streaming attack will incur non-zero processing time.

\paragraph{Relationship to the Ground-Truth Oracle.}
At most operating points, and at every zero-delay configuration, Semantic Gambit surpasses the Ground-Truth (GT) oracle, which keeps the prefix transcript but replaces the LLM forecast with the victim's own transcript of the target window.
Because GT retains the rest of SG's training procedure, the two generators differ only in the content of this channel, so the ordering indicates that a forecast taken from the window under attack is not by itself a stronger training signal than one predicted from the prefix. Together with the AO ablation, which removes the channel entirely, the two comparisons attribute SG's advantage to the presence of a text channel and show that this advantage
does not require the forecast to match what is spoken.

\subsection{Cross-Dataset Transferability}
\label{sec:cross_transfer}

We train generators on LibriSpeech-100 (LS)~\citep{panayotov2015librispeech} and Common Voice 25.0 English (CV)~\citep{ardila2020common_voice} independently under matched compute budgets and evaluate all four train--eval combinations. CV differs substantially from LS in speaker diversity, accent coverage, recording quality, and utterance length (Appendix~\ref{app:cross_dataset_prep}). Because the victim was pretrained on LS, clean WER on CV is 26\% vs.~2\% on LS; we report $\Delta\mathrm{WER}$ (attack WER $-$ clean WER) to control for this baseline gap. The full numerical transferability matrix is in Appendix~\ref{app:cross_full_matrix}.

\begin{figure}[t]
\centering
\includegraphics[width=0.8\textwidth]{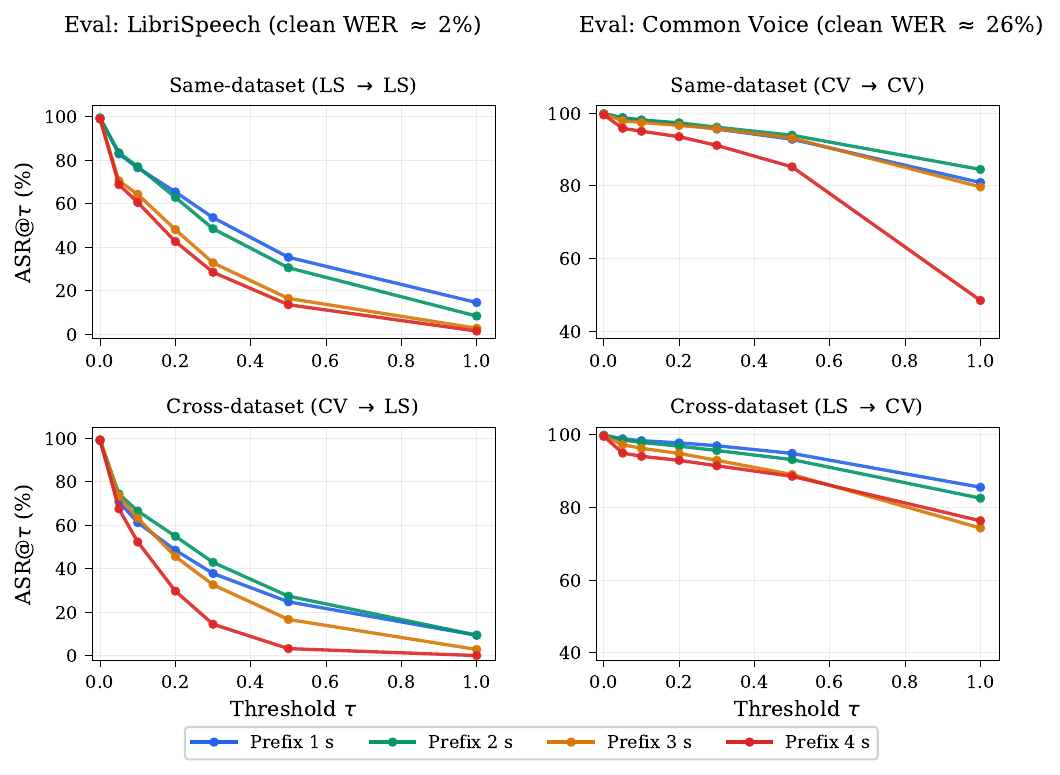}
\caption{ASR@$\tau$ exceedance curves for all four cross-dataset conditions at delay 0.0\,s. Each panel shows four prefix configurations (colors). Top row: same-dataset training; bottom row: cross-dataset training. Left column: evaluation on LibriSpeech; right column: evaluation on Common Voice (\textbf{y-axis 40--100\%, note different scale}). The correspondence between same-dataset and cross-dataset curves %
confirms that transfer preserves the full severity profile, not just attack rates at individual thresholds.}
\label{fig:cross_asr_curves}
\end{figure}

\paragraph{Attack Effectiveness Transfers Across Datasets.}
The generator does not overfit to its training distribution. Figure~\ref{fig:cross_asr_curves} plots ASR@$\tau$ across the full threshold range for all four train--eval combinations. Comparing same-dataset (top row) and cross-dataset (bottom row) panels within each column, the curve shapes are nearly identical at every prefix: the generator produces the same distribution of per-utterance degradation severities regardless of whether it was trained on matched or mismatched data. On LS evaluation at prefixes 2.0--3.0\,s, same-dataset and cross-dataset curves overlap across the full threshold range, with ASR values agreeing within 1 percentage point at both moderate ($\tau{=}0.50$) and severe ($\tau{=}1.00$) thresholds. On CV evaluation the match is tighter still, with curves virtually indistinguishable at prefixes 1.0--3.0\,s. At prefix 4.0\,s the relationship inverts: the cross-dataset generator trained on LS outperforms the matched CV-trained one, demonstrating that a mismatched but acoustically richer training set can surpass a matched but data-constrained one.

\paragraph{Prefix-Length Dependence Generalizes.}
Within each panel of Figure~\ref{fig:cross_asr_curves}, the two shortest prefixes (1.0 and 2.0\,s) dominate the two longest (3.0 and 4.0\,s) across all four train--eval combinations.
This rules out accidental single-prefix transfer and indicates that the generator exploits properties of the victim architecture rather than dataset-specific acoustic patterns: prefix-length difficulty is determined by how the victim processes streaming audio, not by what the generator was trained on.

\subsection{Cross-Model Transferability}
\label{sec:cross_model}
We next ask whether perturbations crafted against one victim ASR model retain effectiveness on an unseen model. Prior work finds that targeted optimization attacks are unlikely to transfer between real-world ASR systems~\citep{Abdullah2022DemystifyingLA}; our attack is untargeted, so its transfer behavior is an open empirical question. We evaluate in a strict black-box transfer setting: the generator is trained against a single CTC surrogate and never updated; at test time, only the transcription model is substituted. We test bidirectional transfer within the CTC self-supervised family (between W2V2 and HuBERT-Large~\citep{hsu2021hubert}) and cross-architecture transfer to Whisper-small~\citep{radford2022whisper}, an encoder--decoder seq2seq model. Table~\ref{tab:cross_model_transfer} reports attack WER at delay 0.0\,s alongside each surrogate's in-domain baseline; the full prefix-by-delay grid is in Appendix~\ref{app:cross_model_transfer}.
\begin{table}[t]
\centering
\small
\caption{Cross-model transfer attack WER (\%) at delay 0.0\,s and 20\,dB SNR. ``In-domain'' is the surrogate evaluated on itself; ``Transfer'' is the same generator evaluated on the other CTC victim. Clean WER $\approx$ 2\% on both models. Full results in Appendix~\ref{app:cross_model_transfer}.}
\label{tab:cross_model_transfer}
\begin{tabular}{l cc cc}
\toprule
\multirow{2}{*}{Prefix}
  & \multicolumn{2}{c}{W2V2 surrogate}
  & \multicolumn{2}{c}{HuBERT surrogate} \\
\cmidrule(lr){2-3} \cmidrule(lr){4-5}
  & In-domain & Transfer & In-domain & Transfer \\
\midrule
1.0\,s & 35.63 & 19.05 & 21.21 & 11.97 \\
2.0\,s & 32.72 & 19.04 & 25.17 & \phantom{0}3.85 \\
3.0\,s & 22.69 & 11.57 & 14.64 & \phantom{0}7.41 \\
4.0\,s & 20.57 & \phantom{0}9.14 & 14.97 & \phantom{0}6.74 \\
\bottomrule
\end{tabular}
\end{table}

\paragraph{Perturbations Transfer Within the CTC Family.}
Adversarial perturbations crafted against one CTC surrogate degrade an unseen CTC victim across every prefix configuration evaluated, with transfer attack WER reaching up to 19\% against a 2\% clean baseline (Appendix~\ref{app:cross_model_transfer:bidirectional}). W2V2 and HuBERT share a raw-waveform CNN frontend, frame-level CTC decoding, and self-supervised pretraining objectives that encourage temporally localized representations, so perturbations exploiting fine-grained frame-level structure in one model find analogous structure in the other. Transfer is asymmetric, however: the W2V2 surrogate produces roughly twice the attack WER of HuBERT across configurations, plausibly because W2V2's contrastive loss encourages linearly separable frame-level features whose adversarial directions generalize more readily, whereas HuBERT's cluster-prediction objective produces more idiosyncratic decision boundaries. Thus, a single CTC surrogate, particularly W2V2, suffices to threaten the broader self-supervised CTC family.%

\paragraph{Transfer Does Not Cross the CTC--seq2seq Boundary.}
Against Whisper-small, 
four black-box attack variants all converge on attack WER within $\sim$1.5 percentage points of the 4\% clean baseline, 
pointing to the victim architecture rather than the attack design as the dominant factor (Appendix~\ref{app:cross_model_transfer:whisper}). Whisper differs from the CTC models on every axis: its autoregressive decoder can override locally corrupted acoustic evidence using surrounding context; its supervised training on 680k hours of labeled audio encourages representations more robust to local distortions than frame-level self-supervised objectives; and its log-mel spectrogram front end occupies a fundamentally different input space from the raw-waveform CNN shared by W2V2 and HuBERT. These results indicate that CTC and seq2seq systems present distinct robustness profiles requiring distinct threat models: perturbations that transfer freely within the CTC family are ineffective across the architectural boundary.

\subsection{Limitations and Future Work}
\label{sec:limitations}
This work's focus was to demonstrate the utility of LLM-guided augmentation in ASR attacks. This work lays the foundations for fertile grounds of new research, including how resilient these attacks are in over-the-air attacks and in the presence of adaptive defenders. Additional performance may be unlocked through updates to the LLM framework employed beyond our choice of Llama~3 8B, especially given our use of 8-bit quantization with fixed token prediction lengths.

\section{Conclusion}

We rethink the information available to adversarial attackers by routing LLM-generated forecasts of upcoming linguistic content into a streaming perturbation generator. Changing the available information in a streaming regime produces results that consistently surpass both acoustic-only streaming baselines and an offline gradient attacker with full waveform access.

Semantic Gambit (SG) improves on the acoustic-only generator by 79\% on average at zero delay, and its gains are largest where the audio bottleneck is tightest. This pattern demonstrates that attack effectiveness scales with the informational content of the conditioning signal. Direct access to the upcoming transcription nevertheless does not yield the strongest attacks, since SG surpasses the ground-truth oracle at most operating points. Observed prefix-length effects persist across datasets and architectures, identifying late-utterance positions as the harder regime. Most crucially, this work establishes that the binding constraint on streaming adversarial attacks is no longer the duration of observed audio but the quality of what can be predicted from it.

\subsection*{AI use statement}

This section discloses how generative AI tools were used in the preparation of this work.

\textbf{What we used generative AI tools for.} We used generative AI tools for the following purposes. (i) Code: to assist in writing and debugging parts of the code that implements our method, the baselines and the evaluation. (ii) Literature discovery: to help search for related work. (iii) Writing: the authors wrote the main text themselves; generative AI tools were used to help revise it for logical flow and fluency, and to help draft parts of the appendix from the authors' code, settings and results. (iv) The manuscript also received automated feedback from an AI-based paper feedback tool during preparation.

\textbf{What we did not use them for.} The scientific contribution of this work is the authors' own. The authors conceived the central idea, that a language model's forecast of upcoming speech can compensate for the causal information bottleneck faced by a real-time attacker, and formulated the information-based taxonomy and the threat model in which it is studied. The authors designed the proposed method, including the generator architecture and the training procedure, and designed the experiments, including the choice of baselines, datasets, operating conditions and ablations. The conclusions drawn from the results are the authors' own. Generative AI tools were not used to conceive the idea, to design the method or the experiments, to generate synthetic datasets, or to formulate or prove mathematical claims.

\textbf{How AI-assisted work was verified.} We have reviewed all AI-assisted work. All AI-assisted code was reviewed, run and tested by the authors. All experiments were run by the authors, and all reported numbers come from these runs. Every work surfaced through AI-assisted search was located and checked by the authors at its original source before being cited. We confirmed that each cited work exists, and checked its authors, title, venue and year against the publisher, DOI or arXiv record. We also checked the cited works against the statements they are cited for. The authors also conducted their own literature review without direct AI assistance. All AI-revised or AI-drafted text was reviewed by the authors, who checked its technical content against the code and the results.

\textbf{Language model as a component of the method.} Separately from the assistance described above, a large language model is part of the proposed attack itself. We use a publicly available pretrained checkpoint (Llama 3 8B), without fine-tuning, for a single purpose: given the partial transcript of the speech heard so far, it forecasts the words likely to follow, and this forecast, together with the partial transcript, is passed to the perturbation generator as a conditioning input. This is a scientific component of the pipeline that we study, not a form of research assistance. The language model plays no other role in this work. In particular, it was not used to create or label evaluation data, to judge or score any output, or to compute any reported metric. Attack effectiveness is measured by word error rate and quantities derived from it, and all reported metrics are computed by scripts with no language model involved.

We take responsibility for the final content of this work, including text, claims or artifacts produced with the aid of generative AI.

\subsection*{Ethics statement}
This work presents methods for generating adversarial attacks against automatic speech recognition (ASR) systems. We acknowledge that such attacks carry a meaningful potential for harm. ASR systems increasingly serve as a primary interface through which people interact with AI, and demonstrating that their outputs can be systematically manipulated raises legitimate safety concerns. These concerns are particularly acute for communities that rely on audio communication over text-based modalities, as such users may have limited ability to independently verify ASR performance or detect adversarial interference. All experiments were conducted offline on public speech corpora against publicly available pretrained models; no deployed system was attacked and no human subjects were involved.

Nevertheless, we subscribe to core tenets of the computer security research community, that conducting adversarial research is crucial to understanding, communicating and (where possible) addressing limitations in current systems. Identifying the risks associated with adversarial manipulation of ASR models requires first establishing what those risks concretely are: under what conditions attacks succeed, what architectural properties they exploit, and how their effectiveness transfers across models and domains. Without this understanding, the research community and downstream deployers of ASR technology operate under a false sense of security, in which vulnerabilities persist but remain unknown. The absence of published attacks does not imply the absence of exploitable vulnerabilities; it implies only that those vulnerabilities are poorly understood. Choosing not to study potential failure modes offers the appearance of robustness without its substance. By systematically exposing and analyzing the conditions under which ASR systems can be reliably manipulated, we aim to provide the empirical foundation upon which more effective defenses and more responsible deployment practices can be built.

\subsection*{Reproducibility statement}

We have aimed to describe the method and the experimental protocol in sufficient detail for reimplementation. The training procedure is given in Algorithm~\ref{alg:semantic_gambit} and the experimental setup in Section~\ref{subsec:setup}, with training hyperparameters and compute in Appendix~\ref{app:training_details} and the implementation of every baseline in Appendix~\ref{app:baseline_details}. The cross-dataset and cross-model protocols are detailed in Appendices~\ref{app:cross_transfer_full} and~\ref{app:cross_model_transfer}. All models used are publicly available pretrained checkpoints, and all experiments use the public LibriSpeech and Common Voice corpora. The funding agreement underpinning this work requires the funding body's approval before any release of the code, including in anonymized form, and this process can only begin once the paper is accepted. We will release the code publicly once that approval is obtained.

\bibliography{refs}
\bibliographystyle{iclr2027_conference}

\appendix
\section{Training and Architectural Details}
\label{app:training_details}

\paragraph{Training Procedure.}
The generator (approximately 7.3M parameters) is trained for 4 epochs with Adam (learning rate $1.5 \times 10^{-4}$, $\beta_1 = 0.9$, $\beta_2 = 0.999$) and an exponential learning rate schedule ($\gamma = 0.99$), decayed once per epoch. No
gradient clipping is applied. Training uses a batch size of 4 on a single NVIDIA A100 80\,GB GPU (or H100 80\,GB for AO runs), with a fixed random seed for reproducibility. Each SG training run (4 epochs) required approximately 15.5 A100 GPU-hours; GT and AO ablations required approximately 3.2 and 1.4 GPU-hours per run, respectively. Across the full $(t_p, \tau)$ grid and both corpora, all reported experiments consumed approximately 500 GPU-hours. The SNR constraint is enforced by rescaling the generator output to the target noise RMS prior to injection; empirical SNR measured on the injected waveform matches the nominal value to within 0.15\,dB throughout training. The final checkpoint after 4 epochs is used for evaluation. The same training procedure applies to the AO ablation and the GT oracle (Section~\ref{subsec:baselines}), differing only in the content of the text channel.

\paragraph{Perturbation Scaling.}
\label{par:perturbation_scaling}
Algorithm~\ref{alg:scale_to_snr} details the \textsc{ScaleToSNR} operator invoked at Line~6 of Algorithm~\ref{alg:semantic_gambit}. The operation applies a single scalar multiplication that modulates only the amplitude of the raw generator output $\tilde\delta$; the waveform shape, which encodes the adversarial content, is produced entirely from causal, prefix-derived inputs by $G_\theta$ (Algorithm~\ref{alg:semantic_gambit}, Line~5). Because the scaling factor is computed under \texttt{torch.no\_grad}, no gradient signal from the target window $x_\star$ reaches the generator during training. 
The rescaled perturbation is injected in the $\tanh$ domain, $x_{\mathrm{adv}} = \tanh(\operatorname{artanh}(x_\star) + \delta)$, so that the perturbed waveform remains within the valid amplitude range $(-1,1)$ by construction. Because $\tanh$ is non-linear, the realized perturbation differs slightly from $\delta$; the empirical SNR measured on $x_{\mathrm{adv}}$ matches the nominal $20\,\mathrm{dB}$ to within $0.15\,\mathrm{dB}$ throughout training.

Since the scaling factor depends on $\mathrm{RMS}(x_\star)$, a natural question is whether this introduces a non-causal dependency at inference time. In a deployment setting the attacker controls the injection gain directly, for instance by calibrating loudspeaker volume or setting a digital mixing coefficient, and therefore does not need to compute the exact RMS of future speech in real time. This observation applies to any causal adversarial audio method: the energy of the yet-unobserved signal is never available at injection time, so every real-time attack must either estimate it or sidestep it through gain calibration. The causal information barrier of Section~\ref{sec:problem_setting} governs the semantic content of the perturbation; amplitude scaling requires only a scalar energy estimate that can be derived causally with negligible loss of constraint precision. Empirically, speech energy is sufficiently stationary that a causal RMS estimate derived from the observed prefix shifts the effective SNR by only 1.3\,dB at the median (mean 1.6\,dB; 90th percentile 3.4\,dB),\footnote{Measured over 1{,}263 LibriSpeech test utterances using equal-length adjacent windows of $3.0$\,s with no delay ($\tau = 0$).} a shift small relative to the 20\,dB budget. We nonetheless use the exact target-window energy in all reported experiments, so that constraint compliance is measured precisely rather than confounded by estimation noise. The perturbation waveform is therefore $\mathcal{F}_{t_p}$-measurable, while the amplitude
normalization uses the exact target-window energy as an evaluation convention.

\begin{algorithm}[t]
\caption{\textbf{\textsc{ScaleToSNR}: Perturbation amplitude
normalization.}}
\label{alg:scale_to_snr}
\begin{algorithmic}[1]
\Require Raw perturbation $\tilde\delta \in \mathbb{R}^{N}$,
         injection-window signal $x_\star \in \mathbb{R}^{N}$,
         target SNR $\mathrm{SNR}_{\text{tgt}}$ (dB)
\Ensure  Constrained perturbation $\delta$ satisfying
         $\mathrm{SNR}(x_\star, \delta) = \mathrm{SNR}_{\text{tgt}}$
\State $\sigma_s \leftarrow \mathrm{RMS}(x_\star)
       = \sqrt{\tfrac{1}{N}\sum_{n=1}^{N} x_\star[n]^2}$
       \Comment{Signal energy}
\State $\sigma_{\text{tgt}} \leftarrow
       \sigma_s \,/\, 10^{\,\mathrm{SNR}_{\text{tgt}}/20}$
       \Comment{Target noise RMS}
\State $\sigma_\delta \leftarrow \mathrm{RMS}(\tilde\delta)$
       \Comment{Current perturbation energy}
\State $\delta \leftarrow
       (\sigma_{\text{tgt}} \,/\, \sigma_\delta)\;\tilde\delta$
       \Comment{Uniform rescaling}
\State \Return $\delta$
\end{algorithmic}
\end{algorithm}

\paragraph{Design Choices.}
The victim model (Wav2Vec 2.0 Large, CTC-finetuned; torchaudio bundle \texttt{WAV2VEC2\_ASR\_LARGE\_LV60K\_960H}) is chosen as a widely benchmarked self-supervised ASR system representative of the CTC family; cross-architecture generality is tested separately in Section~\ref{sec:cross_model}. The language model (Llama 3 8B, 8-bit quantized) balances forecast quality against inference latency within the streaming budget. The 15-token forecast horizon reflects the median token count of a 3.0\,s speech segment in LibriSpeech; generating beyond this yields diminishing returns as excess tokens fall outside the target window. 
The 20,dB SNR constraint follows the distortion measure of~\citet{carlini2018audio_adversarial_examples}, who bound perturbation magnitude in dB relative to the original signal. The 3.0\,s target window is set to match typical ASR chunk sizes in streaming deployments. Ablations comparing the primary generator architecture against three alternative fusion strategies are reported in
Appendix~\ref{app:transformer_architecture}.

\begin{figure*}[t]
\centering
\includegraphics[width=\textwidth]{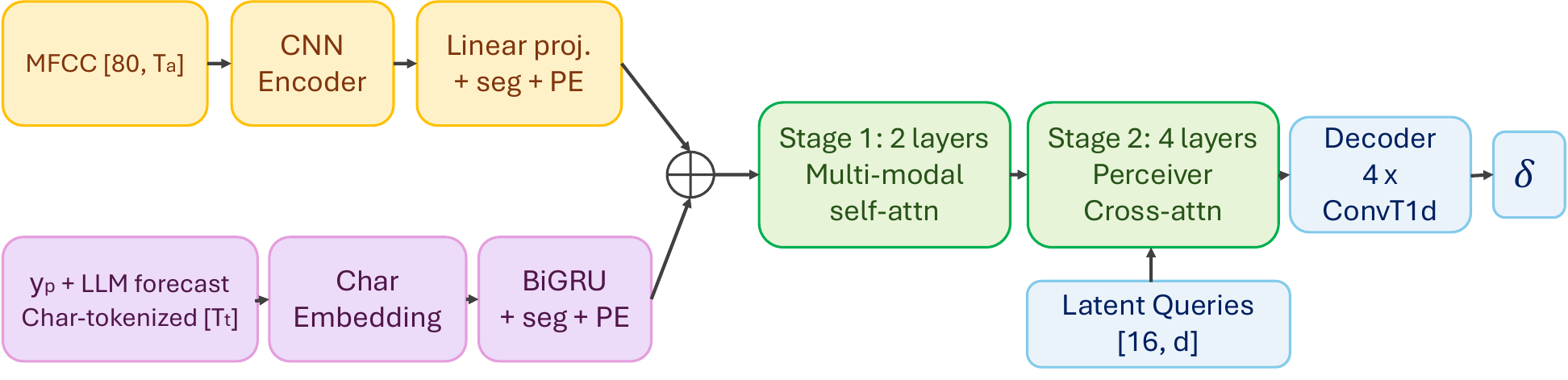}
\caption{Architecture of the SG generator. Audio features (MFCC) and text tokens (prefix transcript + LLM forecast, character-tokenized; see Appendix~\ref{app:training_details}) are encoded separately, annotated with modality and role segment embeddings, and fused through two stages: joint self-attention over the concatenated sequence (Stage 1, 2 layers), followed by Perceiver cross-attention from 16 learnable latent queries (Stage 2, 4 layers). The resulting latent summary is decoded into a waveform perturbation $\delta$ via transposed convolutions. All components are trained jointly.}
\label{fig:sg_arch}
\end{figure*}

\paragraph{Text Tokenization Pipeline.}
The language model $L$ (Llama 3 8B) operates over its native BPE vocabulary and produces up to 15 subword continuation tokens per prefix transcript $y_p$. These tokens are decoded back into a raw string, concatenated with the prefix transcript, and then re-tokenized at the character level using the victim model's CTC alphabet (26 uppercase letters, apostrophe, and word-boundary marker, giving a vocabulary of 29 symbols). The resulting character indices are the input to the generator's text embedding layer (labeled ``Char embedding'' in Figure~\ref{fig:sg_arch}). The 15-token horizon therefore refers to the LLM's generation budget, not to the dimensionality of the generator's text input; the character sequence seen by $G_\theta$ is typically longer than 15 symbols, since each BPE token may decode to multiple characters.

\section{Baseline Implementation Details}
\label{app:baseline_details}

This appendix expands on the baselines summarized in
Section~\ref{subsec:baselines}.

\paragraph{White Noise.}
Gaussian noise is sampled at the input sample rate, applied over
the same temporal extent as adversarial perturbations, and rescaled
per utterance to enforce the SNR budget used by Semantic Gambit. No
optimization is performed.

\paragraph{Acoustic-Only Variants (AO and AO*).}
AO uses the same generator architecture, training procedure, and
hyperparameters as Semantic Gambit but with the text channel
zeroed: both the prefix transcription and the LLM continuation are
replaced with zero vectors at the input to Stage~1, so the
generator conditions only on the audio stream. No other
modifications are made.

AO* replicates the convolutional encoder-decoder of \citet{chiquier2022voice_camouflage}, mapping past spectral
features to future adversarial waveforms without any text input.
Two adaptations are required for comparison under our threat
model: the perturbation budget is enforced via SNR rather than the
original $\ell_\infty$ constraint, and the attack target is
redirected to the Wav2Vec 2.0 victim used throughout this work.
All other architectural details follow the original
implementation. AO* is trained for 4 epochs with Adam (learning
rate $1.5 \times 10^{-4}$) and an exponential schedule
($\gamma = 0.99$); no gradient clipping is applied, following the
original implementation. Together, AO*$\,\to\,$AO$\,\to\,$SG
decomposes improvement over~\citet{chiquier2022voice_camouflage} into an architectural
component (AO*$\,\to\,$AO, information held constant) and an
informational component (AO$\,\to\,$SG, architecture held
constant).

\paragraph{Universal Adversarial Perturbation.}
Universal perturbations for ASR were introduced by~\citet{neekhara2019universal_audio}, who build the perturbation by iteratively aggregating per-utterance minimal perturbations against DeepSpeech. Since their victim and distortion constraint differ from ours, we instead optimize a single 3.0\,s perturbation vector directly, via projected gradient ascent on CTC loss across the LibriSpeech training set.
At inference, the vector is tiled to match the
target utterance length and rescaled per sample to enforce the SNR
constraint. Training uses Adam with a learning rate of $5 \times
10^{-3}$ for 10 epochs; no learning rate schedule or gradient
clipping is applied.

\paragraph{Ground-Truth Oracle (GT).}
The ground-truth forecast $\tilde{y}_\star^{\text{GT}}$ is
obtained by passing the clean target window $x_\star$ through the victim ASR model directly, rather than via forced alignment of the reference transcript. This avoids alignment artifacts and ensures the oracle reflects exactly what the victim would output absent any attack. To match SG's information budget, GT forecasts are truncated to at most 15 tokens using the Llama 3 tokenizer, ensuring the only difference between GT and SG is prediction accuracy rather than token count (denoted with the suffix \texttt{\_tok15}). All other training hyperparameters (Adam,
learning rate $1.5 \times 10^{-4}$, exponential schedule
$\gamma = 0.99$, 4 epochs) follow Semantic Gambit.

\paragraph{Projected Gradient Descent (PGD).}
We use the standard PGD attack of~\citet{madry2018pgd_adversarial_training} with 40 iteration
steps and step size $\alpha = 0.2\,\varepsilon$, where
$\varepsilon$ is the RMS bound implied by the 20\,dB SNR
constraint, following~\citet{chiquier2022voice_camouflage}.
Concretely, the per-utterance perturbation budget is
\begin{equation}
  \varepsilon
  = \lVert x_\star \rVert_{2}\;\cdot 10^{-\mathrm{SNR_{dB}}/20}\enspace,
  \label{eq:snr_budget}
\end{equation}
and the perturbation is updated by
\begin{equation}
  \delta^{(t+1)}
  = \Pi_{\mathcal{B}_{\varepsilon}}\!\biggl(
      \delta^{(t)}
      + \alpha\,
        \frac{\nabla_{\delta}\,
              \mathcal{L}_{\mathrm{CTC}}\!\bigl(
                M(\tilde{x}^{(t)}),\;
                y^{\mathrm{gt}}\bigr)}
             {\bigl\lVert
                \nabla_{\delta}\,
                \mathcal{L}_{\mathrm{CTC}}
              \bigr\rVert_{2}}
    \biggr)\enspace,
  \label{eq:pgd_update}
\end{equation}
where
$\tilde{x}^{(t)}$ denotes the full utterance~$x$ with the target
segment replaced by $x_\star + \delta^{(t)}$,
$\Pi_{\mathcal{B}_{\varepsilon}}$ projects onto
$\{\delta : \lVert\delta\rVert_2 \le \varepsilon\}$,
$M$~is the victim CTC model, and $y^{\mathrm{gt}}$~is the full length ground-truth transcription.
Because PGD is an untargeted attack that maximizes the CTC loss with respect to the ground-truth transcription, the update follows the gradient (ascent) rather than negating it. PGD has non-causal access to the complete utterance during optimization -- the full waveform, including audio beyond the target window, is passed through~$M$ at every iteration -- and so violates the streaming constraint of Section~\ref{sec:problem_setting}; it is included only as a
reference ceiling.

\section{Sentence-Level Results with Confidence Intervals}
\label{app:sentence_level}

\begin{table}[t]
\centering
\caption{\textbf{Sentence-level mean WER (\%)} with 95\% confidence intervals on LibriSpeech under a 20\,dB SNR budget. CIs use the Gaussian approximation $\bar{x} \pm 1.96\,(\mathrm{std}/\sqrt{n})$. Best attack per row in \textbf{bold}. Sample size $n$ varies (859 to 1850) with the prefix, target, and delay window length.}
\label{tab:sentence_ci}
\setlength{\tabcolsep}{8pt}
\renewcommand{\arraystretch}{1.05}
\begin{tabular}{c@{\hskip 10pt}c@{\hskip 14pt}cc@{\hskip 14pt}cc@{\hskip 14pt}cc}
\toprule
\multirow{2}{*}{Prefix} & \multirow{2}{*}{Delay} & \multicolumn{2}{c}{SG (Ours)} & \multicolumn{2}{c}{AO} & \multicolumn{2}{c}{AO$^{*}$} \\
\cmidrule(lr){3-4} \cmidrule(lr){5-6} \cmidrule(lr){7-8}
 & & Mean & 95\% CI & Mean & 95\% CI & Mean & 95\% CI \\
\midrule
\multirow{3}{*}{1.0} & 0.0 & \textbf{49.6} & [47.3, 51.9] & 21.0 & [19.9, 22.2] & 16.4 & [15.6, 17.3] \\
 & 0.5 & 21.9 & [20.7, 23.0] & \textbf{25.8} & [24.5, 27.0] & 10.5 & [9.9, 11.2] \\
 & 1.0 & \textbf{23.7} & [22.6, 24.8] & 18.5 & [17.6, 19.5] & 6.4 & [6.0, 6.9] \\
\midrule
\multirow{3}{*}{2.0} & 0.0 & \textbf{41.9} & [39.9, 43.8] & 15.9 & [15.1, 16.8] & 2.3 & [2.1, 2.5] \\
 & 0.5 & \textbf{25.7} & [24.2, 27.2] & 17.3 & [16.4, 18.1] & 5.1 & [4.8, 5.5] \\
 & 1.0 & \textbf{23.4} & [22.1, 24.6] & 16.5 & [15.5, 17.4] & 3.5 & [3.2, 3.8] \\
\midrule
\multirow{3}{*}{3.0} & 0.0 & \textbf{28.0} & [26.4, 29.6] & 17.9 & [16.9, 18.9] & 4.7 & [4.4, 5.1] \\
 & 0.5 & \textbf{25.4} & [23.9, 26.9] & 23.3 & [22.1, 24.4] & 2.5 & [2.3, 2.7] \\
 & 1.0 & \textbf{13.2} & [12.4, 14.0] & 13.0 & [12.2, 13.7] & 9.7 & [9.1, 10.2] \\
\midrule
\multirow{3}{*}{4.0} & 0.0 & \textbf{24.3} & [22.8, 25.8] & 23.3 & [22.0, 24.7] & 6.5 & [6.1, 6.9] \\
 & 0.5 & \textbf{22.3} & [20.9, 23.8] & 17.4 & [16.4, 18.4] & 5.4 & [5.0, 5.8] \\
 & 1.0 & \textbf{21.5} & [20.3, 22.7] & 20.3 & [19.2, 21.4] & 9.8 & [9.2, 10.3] \\
\bottomrule
\end{tabular}
\end{table}

The corpus-level WER reported in Section~\ref{subsec:results} aggregates errors across all evaluation utterances and is appropriate for assessing overall attack efficacy on the test set as a whole. To complement this view, we report sentence-level mean WER with 95\% confidence intervals (Table~\ref{tab:sentence_ci}), which characterizes the distribution of attack effectiveness across individual utterances rather than the pooled error rate. Confidence intervals use the Gaussian approximation $\bar{x} \pm 1.96\,(\mathrm{std}/\sqrt{n})$, where $n$ ranges from 859 to 1850 depending on the (prefix, delay) configuration. The variation in $n$ reflects the streaming evaluation protocol: at longer prefix and delay combinations, fewer LibriSpeech utterances are long enough to admit a complete (prefix, delay, target) window of the required duration. The clean sentence-level mean WER is stable across configurations at 2.04\% to 2.16\%, within 0.2\% of its corpus-level counterpart, and we omit it from the table.

\paragraph{Statistical Robustness Across the Streaming Grid.}
SG's interval lies strictly above~\citeauthor{chiquier2022voice_camouflage}'s (AO$^{*}$) in every cell, so the improvement over the prior streaming baseline is uniform rather than concentrated in a few favorable configurations. Against the AO ablation, the geometry is more textured: SG's interval sits cleanly above AO's across the short-prefix, zero-delay regime where semantic forecasting is expected to contribute most, and the two methods converge toward statistical parity as prefix length and delay grow. This pattern empirically corroborates the information-hierarchy argument of Section~\ref{subsec:results}, with the separation between methods greatest precisely where the acoustic channel alone is most impoverished. We note one departure at a 1.0\,s prefix and 0.5\,s delay, where AO exceeds SG under non-overlapping intervals; its position at the shortest prefix with non-zero delay suggests a specific interaction with the Perceiver latent bottleneck that warrants separate investigation. \citeauthor{chiquier2022voice_camouflage}'s performance reinforces the information-hierarchy argument from a different angle: even a purpose-built streaming attack collapses to near-clean WER at the tightest configurations when restricted to acoustic information alone.

\paragraph{Sentence- versus Corpus-Level Magnitudes.}
The sentence-level means systematically exceed the corresponding corpus-level WER under attack, and the gap scales with attack strength. At SG's strongest cell (prefix 1.0\,s, delay 0.0\,s), the sentence-level mean of 49.6\% exceeds the corpus-level value of 35.6\% by roughly 14\%; at weaker configurations the gap compresses to a few percent; under clean conditions, where there is no induced error to amplify, the two views agree to within 0.2\%. The discrepancy is methodological. Corpus-level WER weights errors by total reference length, so longer utterances dominate the aggregate, whereas the sentence-level mean treats each utterance equally. Length-stratified analysis confirms the mechanism: at SG's strongest cell, medium-length utterances (6 to 15 words, $n{=}456$) reach a median attack WER of 93.3\%, while long utterances (16 or more words, $n{=}1392$) sit at 27.2\%. The fixed 3.0\,s target window covers a larger fraction of short transcripts, so even a modest number of induced errors translates into a high per-sentence WER, which the equal-weighted mean preserves but the length-weighted aggregate dilutes. We treat both views as complementary: corpus-level WER governs comparisons against the prior streaming literature, which reports the same aggregate, while sentence-level intervals support the per-configuration significance claims above.

\section{Streaming Latency Decomposition}
\label{app:latency}

The injection delay $\tau$ in Section~\ref{subsec:setup} parameterizes the time between the attacker's last observation and the start of the perturbation window. In a deployed system, this delay reflects the wall-clock cost of three sequential operations: (i)~the victim ASR transcribing the observed prefix, (ii)~the LLM generating a semantic forecast from that transcript, and (iii)~the generator producing the adversarial waveform. We measure each component independently and summarize the results in Table~\ref{tab:latency_decomposition}.

\paragraph{Victim ASR Transcription.}
In a streaming deployment, victim inference runs concurrently with audio arrival: the encoder processes frames as they are spoken, so only the final chunk contributes additional delay beyond the prefix duration itself. For CTC-based victims this cost is minimal, since frame-synchronous decoding requires no right-context lookahead. Across all four prefix lengths, Wav2Vec\,2.0 completes its forward pass in under 7\,ms on average, with near-constant latency confirming that this reflects fixed per-call GPU overhead rather than scaling with input duration.

\begin{table}[h]
\centering
\caption{Streaming latency decomposition across the full
$(t_p, \tau)$ grid (NVIDIA A100).}
\label{tab:latency_decomposition}
\begin{tabular}{@{}lccc@{}}
\toprule
Component & Mean (ms) & Range (ms) & Share (\%) \\
\midrule
Victim ASR transcription & 5.9  & 5.22--6.97  & ${<}\,1$ \\
LLM forecast             & 654  & 562--725    & ${\approx}\,99$ \\
Generator forward pass   & 3.3  & 2.56--4.10  & ${<}\,1$ \\
\midrule
Total pipeline           & 663  & 564--729    & 100 \\
\bottomrule
\end{tabular}
\end{table}

\paragraph{LLM Forecasting.}
The LLM forecast (Llama 3 8B, 8-bit quantized, generating up to 15 tokens) is the dominant cost, accounting for about 99\% of the pipeline. Inference averages approximately 654\,ms, ranging from 562 to 725\,ms across the grid. The variation reflects prompt length: longer prefixes produce longer transcripts, yielding a longer LLM context window.

\paragraph{Generator Forward Pass.}
The generator averages 3.3\,ms, ranging from 2.56 to 4.10\,ms. This cost scales mildly with prefix length due to the increased number of encoder frames, but remains negligible in all configurations.

\paragraph{End-to-end Budget.}
The full pipeline completes in 564 to 729\,ms, comfortably within the $\tau = 1.0$\,s budget of our most relaxed configuration. Victim transcription and generator inference together contribute under 12\,ms; the pipeline is bottlenecked entirely by the LLM forecast. This bottleneck is architectural rather than fundamental: replacing Llama 3 8B with a smaller or distilled language model, adopting speculative decoding, or moving to a dedicated inference backend would reduce LLM latency substantially without any change to the attack framework itself. Since the generator and victim ASR components together contribute under 12\,ms, improvements to LLM serving translate directly into tighter real-time budgets. These numbers also align with the latency tolerances of production streaming ASR systems, which typically operate with endpointing latencies of several hundred milliseconds to nearly one second~\citep{li2020towards_fast_and_accurate_streaming}, meaning that an attacker's pipeline cost is comparable to delays already present in deployed systems. Comparison against AO and AO$^{*}$ is uninformative on this axis, since neither invokes a language model; the relevant claim is not that SG is faster than methods that pay no semantic cost, but that the cost it does pay fits within the delays we evaluate.

\section{Transformer Architecture Comparison}
\label{app:transformer_architecture}

This appendix presents the full experimental comparison of four Transformer-based generator variants, all sharing identical front-end encoders and waveform decoder. The variants differ only in the intermediate fusion mechanism that compresses the multi-modal context into a fixed-size latent representation, enabling controlled attribution of performance differences to the fusion strategy alone.

All four variants share a common backbone: audio features and text tokens are first projected into a shared embedding space, then routed through a fusion module that produces a fixed-size latent summary, from which transposed convolutions decode the waveform perturbation $\delta$. The fusion module is built around a Perceiver~\citep{jaegle2021perceiver}, chosen because its learnable latent queries decouple decoder cost from input length, allowing the same architecture to handle variable-length audio and text without quadratic scaling. The variants differ in (i) whether a Stage 1 self-attention encoder precedes the Perceiver, (ii) how modalities are tagged via segment embeddings (2-way audio/text vs.\ 3-way audio/prefix/forecast), and (iii) whether audio and text share a joint softmax inside the Perceiver or are kept as separate streams with gated fusion.

\subsection{Variant Descriptions}

\begin{table}[htbp]
\centering
\caption{Summary of generator architecture variants. All Perceiver modules use 16 latent queries and 4 layers. ``3-way'' denotes segment embeddings for audio, prefix text, and forecast text; ``2-way'' merges the two text segments into one.}
\label{tab:variant_summary}
\begin{tabular}{@{}llllr@{}}
\toprule
\textbf{Variant} & \textbf{Stage 1} & \textbf{Stage 2} & \textbf{Tests} & \textbf{Params} \\ \midrule
Var.\ A & None & Joint softmax, 2-way & 3-way segments + Stage 1 & ${\sim}$5.7M \\
Var.\ B & None & Joint softmax, 3-way & Stage 1 self-attention & ${\sim}$5.7M \\
Var.\ C & None & Gated dual cross-attn & Separated modality streams & ${\sim}$7.5M \\
SG      & 2-layer self-attn & Joint softmax, 3-way & Full model & ${\sim}$7.3M \\ \bottomrule
\end{tabular}
\end{table}

\paragraph{Variant A -- Perceiver Only.} Audio and text tokens are concatenated with two segment embeddings and fused exclusively through a 4-layer Perceiver cross-attention module. The two modalities interact only through the query bottleneck.

\paragraph{Variant B -- Perceiver with 3-way Segment Embeddings.} Identical to Variant A except that text tokens receive per-position segment embeddings distinguishing prefix text (grounded in observed audio) from forecast text (LLM-predicted). Comparing Variant B with SG isolates the contribution of Stage 1, since both share the same segment embedding scheme.

\paragraph{Variant C -- Gated Dual Cross-Attention.} Audio and text are kept as separate streams throughout. At each Perceiver layer, latent queries independently cross-attend to audio and text through separate attention heads, then a learned sigmoid gate combines the outputs per query. Each modality has its own softmax normalization, preventing mutual dilution.

\paragraph{SG -- Multi-Modal Self-Attention + Perceiver.} The full two-stage architecture. Stage 1 consists of 2 TransformerEncoder layers enabling direct bidirectional attention between audio frames and text tokens. Stage 2 is a standard 4-layer Perceiver with 3-way segment embeddings. The additional self-attention layers add ${\sim}$1.6M parameters over Variant A/Variant B.

\subsection{Results and Analysis}
\begin{table*}[!t]
\centering
\small
\caption{Architecture comparison (Target: 3.0\,s): \textbf{Corpus-level} WER (\%) against Wav2Vec 2.0 Large under \textbf{20\,dB SNR}.}
\label{tab:arch_corpus}
\setlength{\tabcolsep}{5pt}
\renewcommand{\arraystretch}{0.95}
\resizebox{0.6\textwidth}{!}{%
\begin{tabular}{@{}cc c ccc c@{}}
\toprule
\multirow{2}{*}{\textbf{Prefix (s)}} & \multirow{2}{*}{\textbf{Delay (s)}} & \multirow{2}{*}{\textbf{Clean}} & \multicolumn{3}{c}{\textbf{Architecture Variants}} & \multirow{2}{*}{\textbf{SG}} \\ \cmidrule(lr){4-6}
 & & & \textbf{Var.\ A} & \textbf{Var.\ B} & \textbf{Var.\ C} & \\ \midrule
\multirow{3}{*}{1.0}
 & 0.0 & 2.05 & 21.84 & 25.18 & 28.31 & 35.63 \\
 & 0.5 & 2.05 & 25.85 & 23.86 & 20.62 & 16.26 \\
 & 1.0 & 2.02 & 17.27 & 17.79 & 21.45 & 18.89 \\
 \addlinespace
\multirow{3}{*}{2.0}
 & 0.0 & 2.02 & 21.20 & 20.82 & 11.69 & 32.72 \\
 & 0.5 & 2.01 & 21.97 & 20.86 & 14.12 & 20.32 \\
 & 1.0 & 2.03 & 16.62 & 28.28 & 14.70 & 18.95 \\
 \addlinespace
\multirow{3}{*}{3.0}
 & 0.0 & 2.03 & 14.96 & 16.63 & 16.44 & 22.69 \\
 & 0.5 & 2.02 & 11.48 & 24.13 & 14.00 & 21.17 \\
 & 1.0 & 2.01 & 20.91 & 12.35 & 21.47 & 11.32 \\
 \addlinespace
\multirow{3}{*}{4.0}
 & 0.0 & 2.01 & 15.12 & 17.57 & 15.24 & 20.57 \\
 & 0.5 & 1.98 & 24.66 & 18.96 & 13.03 & 19.12 \\
 & 1.0 & 1.97 & 12.84 & 11.94 & 15.38 & 18.68 \\ \bottomrule
\end{tabular}%
}
\end{table*}

\begin{table*}[!t]
\centering
\small
\caption{Architecture comparison (Target: 3.0\,s): \textbf{Sentence-level mean} WER (\%) against Wav2Vec 2.0 Large under \textbf{20\,dB SNR}.}
\label{tab:arch_mean}
\setlength{\tabcolsep}{5pt}
\renewcommand{\arraystretch}{0.95}
\resizebox{0.6\textwidth}{!}{%
\begin{tabular}{@{}cc c ccc c@{}}
\toprule
\multirow{2}{*}{\textbf{Prefix (s)}} & \multirow{2}{*}{\textbf{Delay (s)}} & \multirow{2}{*}{\textbf{Clean}} & \multicolumn{3}{c}{\textbf{Architecture Variants}} & \multirow{2}{*}{\textbf{SG}} \\ \cmidrule(lr){4-6}
 & & & \textbf{Var.\ A} & \textbf{Var.\ B} & \textbf{Var.\ C} & \\ \midrule
\multirow{3}{*}{1.0}
 & 0.0 & 2.16 & 30.46 & 34.58 & 38.68 & 49.60 \\
 & 0.5 & 2.15 & 34.68 & 32.19 & 27.71 & 21.90 \\
 & 1.0 & 2.08 & 21.98 & 22.36 & 27.61 & 23.72 \\
 \addlinespace
\multirow{3}{*}{2.0}
 & 0.0 & 2.08 & 27.30 & 26.38 & 14.92 & 41.86 \\
 & 0.5 & 2.09 & 27.17 & 25.72 & 17.38 & 25.69 \\
 & 1.0 & 2.12 & 20.08 & 35.11 & 17.89 & 23.35 \\
 \addlinespace
\multirow{3}{*}{3.0}
 & 0.0 & 2.12 & 18.33 & 20.39 & 19.87 & 28.01 \\
 & 0.5 & 2.12 & 13.59 & 28.81 & 16.83 & 25.44 \\
 & 1.0 & 2.10 & 24.55 & 14.25 & 25.55 & 13.18 \\
 \addlinespace
\multirow{3}{*}{4.0}
 & 0.0 & 2.10 & 17.67 & 20.67 & 17.79 & 24.33 \\
 & 0.5 & 2.06 & 28.82 & 22.04 & 16.26 & 22.33 \\
 & 1.0 & 2.04 & 14.60 & 13.55 & 17.74 & 21.50 \\ \bottomrule
\end{tabular}%
}
\end{table*}

At zero delay, SG achieves the highest corpus-level WER at every prefix length, with a substantial margin at short prefixes (35.63\% vs the next-best 28.31\% from Variant C at prefix of 1.0\,s). The advantage narrows at longer prefixes (20.57\% vs 17.57\% from Variant B at prefix of 4.0\,s), consistent with the general observation that all text-containing architectures converge toward similar performance as prefix length increases.

\paragraph{Contribution of Stage 1 (Variant B vs SG).}
Comparing Variant B and SG isolates the effect of the 2-layer self-attention encoder. At zero delay, SG consistently outperforms Variant B: by 10.45 percentage points at prefix of 1.0\,s, 11.90 percentage points at prefix of 2.0\,s, 6.06 percentage points at prefix of 3.0\,s, and 3.00 percentage points at prefix of 4.0\,s. The diminishing gap at longer prefixes suggests that Stage 1's cross-modal attention is most valuable when the context is compact enough for the model to learn meaningful audio--text correspondences.

\paragraph{Value of Segment Embeddings (Variant A vs Variant B).}
Variant A and Variant B differ only in their segment embedding scheme: 2-way vs 3-way. At zero delay, Variant B outperforms Variant A at three of four prefix lengths (by 3.34 percentage points at prefix of 1.0\,s, 1.67 percentage points at prefix of 3.0\,s, and 2.45 percentage points at prefix of 4.0\,s), while Variant A holds a marginal edge at prefix of 2.0\,s (21.20\% vs 20.82\%). The benefit of distinguishing prefix transcription from LLM forecast is therefore modest but broadly consistent.

\paragraph{Separate Modality Streams (Variant C vs SG).}
Variant C's gated dual cross-attention underperforms SG at zero delay across all prefix lengths (28.31\% vs 35.63\% at prefix of 1.0\,s; 15.24\% vs 20.57\% at prefix of 4.0\,s). Variant C tests whether a more powerful separate-stream mechanism (gated dual cross-attention with dedicated per-modality softmax normalization and ${\sim}$7.5M parameters, exceeding even SG's ${\sim}$7.3M) can match joint self-attention. The fact that it cannot suggests that the disadvantage of separate modality streams is not purely a capacity limitation but reflects a more fundamental issue: keeping audio and text in isolated pathways, even with learned gating, prevents the fine-grained token-level interactions that joint self-attention enables in SG. The separate-stream design may preserve cleaner per-modality representations, but at the cost of the direct audio--text alignment that appears critical for generating effective perturbations.

However, Variant C shows more competitive results at a delay of 1.0\,s (e.g.\ 21.45\% and 21.47\% at prefix of 1.0\,s and 3.0\,s), suggesting that the separate-stream design may have advantages in the delayed-injection scenario where temporal alignment between modalities is less direct. More broadly, the configuration-dependent instability observed at non-zero delays persists across all four Transformer variants, which suggests that this variance is a systematic property of adversarial CTC optimization rather than an artifact of any particular fusion design.

\section{Investigating the Prefix Length Effect}
\label{sec:appendix_prefix}

The corpus-level WER results in the main paper (Table~\ref{tab:baselines_corpus_20db}) show that the WER of all text-conditioned attacks decreases as the observed prefix grows longer at zero delay, while the acoustic-only ablation does not.
The same pattern holds across every text-containing generator variant we evaluated, which rules out the fusion architecture as the cause: Variant A, Variant B, Variant C and SG are all lower at a 4.0\,s prefix than at 1.0\,s (Appendix~\ref{app:transformer_architecture}, Tables~\ref{tab:arch_corpus}--\ref{tab:arch_mean}).
This appendix documents the investigation behind the main paper's claim that the decline reflects an intrinsic diminishing of the text channel's marginal value. We test three hypotheses, conduct attention-map analysis, run two architectural interventions, and rule out training-budget confounds; the effect does not track any specific architectural design choice, and is consistent with a property of the prediction problem itself.

\subsection{The Observation}
The main paper attributes the prefix-length decline to two compounding mechanisms: a victim-side effect, in which longer prefixes give the ASR a cleaner anchor for decoding, and a text-channel effect specific to text-conditioned attacks. The decline is broad: PGD, AO\textsuperscript{*}, GT, and SG all decline with prefix length, but its magnitude varies across methods, and the contrast that frames this appendix's investigation is the gap between text-conditioned attacks and the acoustic-only ablation. If the victim-side effect were the dominant mechanism, the acoustic-only ablation should decline at a comparable rate to the text-conditioned variants. It does not (corpus-level WER of 15.15\%, 12.58\%, 14.82\%, 19.85\% across the four prefix lengths), suggesting that an additional, text-channel-specific mechanism is contributing to the steeper declines observed in Variant A, Variant B, SG, and Variant C.
Quantitatively, SG's advantage over the acoustic-only ablation collapses from +20.48 points at a prefix of 1.0\,s to just +0.72 points at a prefix of 4.0\,s. The text channel's contribution to attack effectiveness diminishes sharply as prefix length grows.

\subsection{Hypotheses and Interventions}

Three hypotheses were formulated and tested.

\paragraph{Hypothesis 1: Victim-Side CTC Anchor Effect.}
A longer clean prefix provides the victim ASR with more correctly-aligned frames, creating a stronger alignment anchor that helps resist corruption in the target window. The GT oracle declines from 23.53\% to 12.44\% across prefix lengths, confirming that the attack problem does become harder. This is unlikely to be the sole explanation, however, since the acoustic-only ablation, which faces the same victim, does not decline.

\paragraph{Hypothesis 2: Perceiver Bottleneck.}
The fixed 16 latent queries may be insufficient to compress the growing multi-modal context (108 tokens at prefix\,=\,1.0\,s vs 168 at prefix\,=\,4.0\,s). Attention map analysis confirmed that attention entropy saturates at longer prefixes: peak weights collapse from 0.05--0.20 at prefix\,=\,1.0\,s to 0.008--0.018 at prefix\,=\,4.0\,s, and Shannon entropy approaches the theoretical maximum (see Subsection~\ref{subsec:attention_analysis_detail}). We tested two interventions (Table~\ref{tab:interventions}):

\begin{table}[htbp]
\centering
\caption{Architectural interventions targeting the Perceiver bottleneck, compared against standard SG (delay\,=\,0.0\,s, 20\,dB SNR).}
\label{tab:interventions}
\begin{tabular}{@{}c ccc@{}}
\toprule
\textbf{Prefix (s)} & \textbf{SG} & \textbf{32-query} & \textbf{Sep-softmax} \\ \midrule
1.0 & 35.63 & 19.70 & 24.28 \\
2.0 & 32.72 & 13.75 & 19.39 \\
3.0 & 22.69 & 19.79 & 16.64 \\
4.0 & 20.57 & 21.28 & 14.65 \\ \bottomrule
\end{tabular}
\end{table}

\begin{itemize}
    \item \textbf{Increased query count ($n_{\text{latent}} = 32$):} 
    Performance degraded at three of the four prefix lengths (e.g.\ 19.70\% vs 35.63\% at prefix\,=\,1.0\,s) and was marginally higher at prefix\,=\,4.0\,s (21.28\% vs 20.57\%). However, this result is confounded by the waveform decoder's architectural coupling to 16 temporal frames; doubling the query count disrupts the decoder's transposed-convolution chain, making the comparison inconclusive.
    \item \textbf{Separate-softmax cross-attention:} Audio and text tokens receive independent softmax normalization with a learned gate combining the outputs. Performance degraded at all four prefix lengths, including short prefixes where entropy saturation was not observed (24.28\% vs 35.63\% at prefix\,=\,1.0\,s, and 14.65\% vs 20.57\% at 4.0\,s).
    This demonstrates that the joint softmax is actively beneficial, not a bottleneck.
\end{itemize}

Taken together, the entropy saturation observed in the attention analysis is a real correlational phenomenon but is not causally responsible for the WER decline. The joint softmax and fixed query count are well-functioning components of the architecture.

\paragraph{Hypothesis 3: Diminishing Marginal Value of Text.}
The text channel's informational value for generating adversarial perturbations may inherently diminish at longer prefixes. The text channel's contribution (SG minus acoustic-only) collapses from ${\sim}$20 points to $<$1 point. The acoustic-only ablation, which processes only audio tokens (13--51 tokens across prefix lengths) and never faces the growing text context, does not decline. Since the decline is universal across architectures with fundamentally different fusion mechanisms (joint attention, separate streams, with/without Stage 1), it cannot be attributed to any particular design choice.
This explanation is consistent with the evidence above, and we do not separate it from the victim-side effect of Hypothesis 1.

\subsection{Training Convergence}

One potential confound is that longer-prefix models, which process more context tokens per step, may be under-trained given the same training budget (4 epochs, ${\sim}$28,500 steps). We rule this out by examining the final training CTC loss: at both prefix\,=\,1.0\,s and prefix\,=\,4.0\,s, the loss in the final 1,500 steps oscillates in the same range (approximately 0.2--2.5) with comparable gradient norms and identical learning rates. Neither model shows a systematic downward trend, indicating that both have converged.

\subsection{Attention Map Analysis}
\label{subsec:attention_analysis_detail}

To understand the internal dynamics of SG under varying prefix configurations, we extracted attention weights from all four Perceiver decoder layers by patching PyTorch's \texttt{MultiheadAttention} modules to capture per-head matrices during evaluation. For each prefix configuration (1.0--4.0\,s, delay\,=\,0.0\,s), we analyzed 20 batches (${\sim}$60 samples).

\paragraph{Modality Balance.}
The audio attention fraction remains stable at 30--40\% across all prefix configurations (Table~\ref{tab:attn_fractions_appendix}). The Pearson correlation with WER is weak ($r = -0.29$), ruling out the hypothesis that audio tokens progressively dominate the softmax at longer prefixes. The same picture holds at the per-query level: audio fractions fall in the 15--60\% range at both prefix lengths, with no query attending exclusively to one modality and the aggregate balance broadly preserved.

\begin{table}[htbp]
\centering
\caption{Mean audio attention fraction (\%) in Stage\,2 cross-attention by decoder layer (delay\,=\,0.0\,s, 20\,dB SNR).}
\label{tab:attn_fractions_appendix}
\begin{tabular}{@{}c cccc c@{}}
\toprule
\textbf{Prefix (s)} & \textbf{L0} & \textbf{L1} & \textbf{L2} & \textbf{L3} & \textbf{Avg} \\ \midrule
1.0 & 33.8 & 37.0 & 32.2 & 46.8 & 37.5 \\
2.0 & 38.5 & 33.2 & 29.3 & 17.1 & 29.5 \\
3.0 & 45.1 & 52.1 & 29.6 & 35.7 & 40.6 \\
4.0 & 35.3 & 33.5 & 37.3 & 30.6 & 34.2 \\ \bottomrule
\end{tabular}
\end{table}

\paragraph{Peak Weight Collapse and Entropy Saturation.}
While modality balance is broadly preserved, attention selectivity is not. For a representative example, peak attention weight at the final Perceiver layer collapses roughly $10\times$ from prefix $= 1.0\,\mathrm{s}$ (peaks of 0.15--0.35) to prefix $= 4.0\,\mathrm{s}$ (peaks barely exceeding 0.02). Shannon entropy quantifies the same loss of selectivity: at prefix $= 1.0\,\mathrm{s}$ (80 total tokens), query entropies sit in the 3.5--4.0 bit range against a theoretical maximum of 6.3 bits, leaving substantial headroom for selective attention; at prefix $= 4.0\,\mathrm{s}$ (170 total tokens), entropies push to roughly 7.0 bits against a maximum of 7.4 bits, within 0.5 bits of saturation.

\subsection{Summary}

The prefix length effect is a robust empirical phenomenon that cannot be attributed to any specific architectural design choice, training insufficiency, or attention mechanism failure. 
The evidence is consistent with a diminishing marginal value of semantic conditioning as the prefix length increases, compounded with the victim-side anchoring effect. 
At short prefixes, the text channel provides substantial additional attack capability (+20 points over acoustic-only); at long prefixes, this advantage is nearly eliminated. 
That the acoustic-only ablation does not decline indicates that a text-channel-specific component contributes beyond the general increase in difficulty.

\section{Cross-Dataset Transferability}
\label{app:cross_transfer_full}

This appendix provides the complete experimental results supporting Section~\ref{sec:cross_transfer}. Section~\ref{app:cross_dataset_prep} documents the Common Voice preparation pipeline and summarizes the key differences between the two datasets. Section~\ref{app:cross_training} details the matched-compute training configuration. Section~\ref{app:cross_full_matrix} reports the full $2\times 2\times 4$ transferability matrix (two training datasets $\times$ two evaluation datasets $\times$ four prefix configurations). Section~\ref{app:cross_matched_subset} addresses the subset-matching confound introduced by prefix-dependent duration filters. Section~\ref{app:cross_length_strat} reports length-stratified results. Section~\ref{app:cross_constraint} documents constraint compliance. All experiments in this appendix use zero injection delay, so that the dataset is the only factor varied; clips are retained only if they cover the prefix and the full 3.0\,s target window.

\subsection{Common Voice Preparation}
\label{app:cross_dataset_prep}

Common Voice 25.0 English was obtained from the Mozilla Data Collective (87.84\,GiB, 2.57M clips across train/dev/test splits). The official test split (16{,}398 clips) was retained in full. The train split (1{,}147{,}465 validated clips) was randomly sub-sampled to 250{,}000 clips. MP3 clips were decoded with librosa, resampled to 16\,kHz mono, filtered to a minimum duration of 4.0\,s (to accommodate our loosest prefix configuration, prefix 1.0\,s plus target 3.0\,s), and written as uncompressed WAV. Audio-decode failures affected 5.7\% of clips and were dropped. Text transcripts were normalized to match LibriSpeech conventions: uppercased, punctuation stripped, whitespace collapsed, apostrophes retained. After preprocessing and downsampling, the effective training set contained 250{,}000 clips and the test set 13{,}606 clips passing the duration filter (longer-prefix configurations evaluate on smaller subsets; see Section~\ref{app:cross_matched_subset}).

Table~\ref{tab:dataset_comparison} summarizes the key differences between the two datasets. These differences make the cross-dataset experiment a stringent test of generalization: a generator trained on one distribution must produce perturbations that degrade the victim on audio with substantially different acoustic properties.

\begin{table}[h]
\centering
\small
\caption{Summary of dataset differences between LibriSpeech-100 and Common Voice 25.0 English.}
\label{tab:dataset_comparison}
\begin{tabular}{lcc}
\toprule
 & LibriSpeech-100 & Common Voice \\
\midrule
Speakers       & $\sim$1{,}000     & 99{,}724 \\
Accent coverage & US English (audiobook) & Global English \\
Recording conditions & Studio, read speech & Consumer mic, varied \\
Mean clip length & 12.7\,s & 5.3\,s \\
Training clips   & 28{,}539 & 250{,}000 \\
Test clips (prefix 1.0\,s) & 1{,}850 & 13{,}606 \\
Clean WER (W2V2) & $\approx$2\% & $\approx$26\% \\
\bottomrule
\end{tabular}
\end{table}

\subsection{Training Configuration}
\label{app:cross_training}

Each prefix configuration (1.0\,s, 2.0\,s, 3.0\,s, 4.0\,s; delay 0\,s, target 3\,s) was trained for 28{,}500 gradient steps at batch size 4, learning rate $1.5\times 10^{-4}$. This exactly matches the step budget used for the LS-100 experiments, yielding 114{,}000 clip-passes per run. The LS-100 pool (28{,}539 clips) is smaller than the CV subsample (250{,}000 clips), so LS training completes approximately 4 epochs while CV training completes approximately 0.45 epochs. 
Same-dataset attack rates differ between the two regimes: CV training reaches ASR@0.10 of 95.0 to 98.1\% on CV, while LS training reaches 60.6 to 77.0\% on LS (Table~\ref{tab:cross_full}). Since the two victims' clean error rates also differ substantially (26\% against 2\%), these rates are not directly comparable, and we match compute rather than epochs; the residual effect of the difference in sample exposure (approximately 4 epochs against 0.45) is not isolated by our design.

All other hyperparameters are identical to the main paper: SG generator with 16 latent queries, joint softmax, SNR 20\,dB constraint, $k=15$ forecast tokens from Meta-Llama-3-8B. Each run consumed approximately 4 GPU-hours on an NVIDIA H100.

\subsection{Full Transferability Matrix}
\label{app:cross_full_matrix}

Table~\ref{tab:cross_full} reports all sixteen cells of the transferability matrix with evaluation-set sizes and clean WER baselines. Each row is one $(\mathrm{train},\mathrm{eval},\mathrm{prefix})$ combination. Evaluation subsets decrease with prefix length due to the duration filter.

\begin{table}[h]
\centering
\small
\caption{Full cross-dataset transferability matrix across all four prefix configurations. 
$N$ is the number of evaluation utterances long enough to cover the prefix and the 3.0\,s target window. $\Delta\mathrm{WER}$ is the median per-utterance WER increase in percentage points. ASR@$\tau$ is the fraction of utterances with $\Delta\mathrm{WER} \geq \tau$ (expressed as a percentage).
Clean WER is constant within each (eval, prefix) pair and is reported once per group.}
\label{tab:cross_full}
\begin{tabular}{llrrrrrr}
\toprule
Train & Eval & Prefix & $N$ & Clean WER & $\Delta\mathrm{WER}$ (\%) & ASR@0.10 & ASR@1.00 \\
\midrule
\multirow{4}{*}{LS} & \multirow{4}{*}{LS}
  & 1.0 & 1{,}850 & 2.05\% & $+33$ & 76.5\% & 14.6\% \\
& & 2.0 & 1{,}531 & 2.02\% & $+29$ & 77.0\% & 8.4\% \\
& & 3.0 & 1{,}263 & 2.03\% & $+19$ & 64.3\% & 2.8\% \\
& & 4.0 & 1{,}051 & 2.01\% & $+16$ & 60.6\% & 1.5\% \\
\midrule
\multirow{4}{*}{LS} & \multirow{4}{*}{CV}
  & 1.0 & 13{,}606 & 25.76\% & $+183$ & 98.3\% & 85.5\% \\
& & 2.0 & 10{,}826 & 26.06\% & $+182$ & 97.8\% & 82.5\% \\
& & 3.0 & 7{,}530  & 26.54\% & $+155$ & 96.2\% & 74.3\% \\
& & 4.0 & 4{,}697  & 27.39\% & $+158$ & 94.0\% & 76.3\% \\
\midrule
\multirow{4}{*}{CV} & \multirow{4}{*}{LS}
  & 1.0 & 1{,}850 & 2.05\% & $+18$ & 61.2\% & 9.4\% \\
& & 2.0 & 1{,}531 & 2.02\% & $+24$ & 66.5\% & 9.5\% \\
& & 3.0 & 1{,}263 & 2.03\% & $+17$ & 63.2\% & 2.9\% \\
& & 4.0 & 1{,}051 & 2.01\% & $+11$ & 52.3\% & 0.1\% \\
\midrule
\multirow{4}{*}{CV} & \multirow{4}{*}{CV}
  & 1.0 & 13{,}606 & 25.76\% & $+175$ & 98.0\% & 80.9\% \\
& & 2.0 & 10{,}826 & 26.06\% & $+192$ & 98.1\% & 84.5\% \\
& & 3.0 & 7{,}530  & 26.54\% & $+157$ & 97.4\% & 79.7\% \\
& & 4.0 & 4{,}697  & 27.39\% & $+94$  & 95.0\% & 48.5\% \\
\bottomrule
\end{tabular}
\end{table}

Two observations beyond those in the main text. First, evaluation-set sizes shrink substantially with prefix length, so longer-prefix results are computed on a progressively smaller (and longer-utterance) subset; the implications of this subset dependence are discussed in Section~\ref{app:cross_matched_subset}. 
Second, at the longest prefix, the CV-trained generator shows a pronounced drop in attack effectiveness when evaluated on CV (ASR@1.00 of 48.5\% against 79.7\% at prefix 3.0\,s), while the LS-trained generator does not. CV's shorter mean clip length (5.3\,s) leaves fewer training examples that can accommodate a 4.0\,s prefix alongside a full target window; we do not isolate this factor experimentally.

\paragraph{Two Attack Regimes.}
The exceedance curves in Figure~\ref{fig:cross_asr_curves} also reveal that the two evaluation datasets occupy qualitatively different attack regimes. On LS, curves are concave with a steep initial drop and a long tail, reflecting a spread of degradation severities from mild to severe. On CV, curves plateau above 80\% until $\tau \approx 0.5$ before dropping, indicating near-universal disruption up to moderate severity. These correspond to partial-disruption and total-disruption regimes, respectively, driven by the victim's baseline accuracy on each dataset. Crucially, transfer operates in both: the generator trained on mismatched data reproduces both the concave LS profile and the flat CV plateau.

\subsection{Subset-Matching Caveat}
\label{app:cross_matched_subset}

Each prefix configuration evaluates on the subset of the test set whose clips are long enough to cover the prefix and the target window. Direct numerical comparison across prefix configurations therefore conflates attack effectiveness with evaluation-subset characteristics (e.g., longer clips may contain more syntactic context that the attack must disrupt). The same filter governs the prefix rows of Table~\ref{tab:baselines_corpus_20db} and
the prefix-length analysis of Appendix~\ref{sec:appendix_prefix}: at delay 0.0\,s the four
prefix rows use the same four subsets as the LS rows of Table~\ref{tab:cross_full}
($N = 1{,}850, 1{,}531, 1{,}263, 1{,}051$).

The qualitative ranking across prefixes is the same in the LS-only results of Section~\ref{sec:appendix_prefix}, which are subject to the same subset dependence. For cross-dataset comparison at a fixed prefix and a fixed evaluation set (the primary claim of Section~\ref{sec:cross_transfer}), this confound does not apply: both same-dataset and cross-dataset generators at a given prefix are evaluated on identical subsets.

\subsection{Length-Stratified Results}
\label{app:cross_length_strat}
 
Does the attack affect short utterances differently from long ones? This matters for two reasons. First, short utterances mechanically inflate $\Delta\mathrm{WER}$ because WER normalizes by reference length--a handful of hallucinated tokens against a 3-word reference produces a $\Delta\mathrm{WER}$ several times larger than the same hallucination against a 20-word reference. ASR@0.10 sidesteps this inflation because it only asks whether the attack raised WER by at least 10 percentage points. Second, longer utterances give the victim more contextual information for partial recovery, so length-sensitivity reveals whether the attack succeeds by local corruption or by systematic transcript disruption.
 
We partition test utterances by reference-transcript word count (Short: 1--5 words; Medium: 6--15 words; Long: 16+ words) and report ASR@0.10 within each bin at prefix 2.0\,s. We select this configuration because it avoids the extremes of the prefix range: prefix 1.0\,s retains the largest evaluation subset and may overrepresent short CV clips, while prefix 4.0\,s retains the smallest and skews toward longer utterances. Prefix 2.0\,s provides a more balanced view of the length distribution across both datasets.
 
\begin{table}[h]
\centering
\small
\caption{Length-stratified attack success (ASR@0.10) at prefix 2.0\,s. Cell values are the fraction of utterances where the attack raised WER by at least 10 percentage points. The Short bin is empty for LS evaluation because LS test-clean (audiobook narration) contains no clips with transcripts of fewer than six words.}
\label{tab:cross_length}
\begin{tabular}{llrrr}
\toprule
Train & Eval & Short (1--5) & Medium (6--15) & Long (16+) \\
\midrule
LS & LS & -- & 91.2\% ($n$=194)    & 74.9\% ($n$=1{,}337) \\
LS & CV & 95.3\% ($n$=383) & 97.9\% ($n$=10{,}189) & 95.7\% ($n$=254) \\
CV & LS & -- & 89.7\% ($n$=194)    & 63.1\% ($n$=1{,}337) \\
CV & CV & 95.3\% ($n$=383) & 98.2\% ($n$=10{,}189) & 96.5\% ($n$=254) \\
\bottomrule
\end{tabular}
\end{table}
 
Two patterns emerge. On CV evaluation, attack success is near-universal across all length bins (95--98\%): the victim's 26\% clean WER leaves so little baseline performance that even long utterances offer no recovery. On LS evaluation, success drops noticeably from Medium to Long bins--91.2\% to 74.9\% for LS to LS and 89.7\% to 63.1\% for CV to LS--indicating that a well-performing victim can use longer context to partially recover from the perturbation. This length-sensitivity is consistent with the victim-difficulty explanation discussed in Section~\ref{sec:cross_transfer}: the attack's apparent weakness on LS is bounded by how little room the victim has to degrade, and part of that remaining room is absorbed by contextual redundancy in longer utterances.

\subsection{Constraint Compliance}
\label{app:cross_constraint}

Table~\ref{tab:cross_constraint} documents empirical SNR and $L_\infty$ perturbation magnitudes across all four matrix cells at prefix 2.0\,s. Empirical SNR is tightly clustered around the 20\,dB target in every condition, confirming that the generator respects the perturbation budget and that the effectiveness differences reported above are not driven by constraint violations. 

\begin{table}[h]
\centering
\small
\caption{Empirical SNR (dB) and $L_\infty$ perturbation magnitudes at prefix 2.0\,s. Mean $\pm$ standard deviation over the full evaluation set. Target SNR was 20\,dB.}
\label{tab:cross_constraint}
\begin{tabular}{llcc}
\toprule
Train & Eval & SNR (dB) & $L_\infty$ \\
\midrule
LS & LS & $20.03 \pm 0.02$ & $0.034 \pm 0.011$ \\
LS & CV & $20.08 \pm 0.17$ & $0.044 \pm 0.035$ \\
CV & LS & $20.03 \pm 0.02$ & $0.049 \pm 0.016$ \\
CV & CV & $20.08 \pm 0.17$ & $0.063 \pm 0.049$ \\
\bottomrule
\end{tabular}
\end{table}

\section{Cross-Model Transferability}
\label{app:cross_model_transfer}

We assess whether adversarial perturbations crafted by our attack retain effectiveness when transcribed by an unseen victim ASR model. This is a strict black-box transfer setting: the generator is trained against one surrogate and never updated; at evaluation time we substitute only the model that produces the final transcription. The MFCC frontend, prefix transcription, LLM forecasting, and token encoding paths all remain aligned with the training-time surrogate, since these are part of the attack pipeline rather than the victim.

We evaluate three transfer scenarios spanning two architectural families. Within the CTC self-supervised family, we measure transfer in both directions of the W2V2 and HuBERT pair: (i) W2V2-trained generators transcribed by HuBERT-Large-LS960-ft~\citep{hsu2021hubert}, and (ii) HuBERT-trained generators transcribed by W2V2. To probe cross-family transfer, we additionally evaluate (iii) W2V2-trained generators transcribed by Whisper-small~\citep{radford2022whisper}, an encoder--decoder seq2seq model trained with weak supervision on 680k hours of multilingual audio.

\subsection{Setup}
\label{app:cross_model_transfer:setup}

For each (prefix, delay) configuration and each surrogate, we load the generator checkpoint trained against that surrogate, generate adversarial audio for the LibriSpeech \texttt{test-clean} split, and transcribe both clean and adversarial waveforms through the transfer victim. Empirical SNR is held at 20\,dB across all runs (median 20.03\,dB, $\sigma < 0.03$\,dB), confirming that the perturbation budget is preserved end-to-end. The number of evaluable utterances varies with the total audio length required (prefix\,+\,delay\,+\,target $\leq T$), ranging from 1{,}850 utterances for the shortest configuration to 859 for the longest. Clean WER on each victim is approximately 2.0\% across configurations.

We report corpus-level WER on the full utterance and target-window WER restricted to the perturbed segment, which more directly measures local attack efficacy. We additionally report the ASR@$\tau$ family of attack-success metrics, defined as the fraction of utterances with $\Delta\mathrm{WER} \geq \tau$.

\subsection{HuBERT-Surrogate In-Domain Effectiveness}
\label{app:cross_model_transfer:hubert_indomain}

We first establish that the HuBERT surrogate produces a viable in-domain attack across the full prefix$\times$delay grid, mirroring the W2V2-surrogate baseline shown in Table~\ref{tab:baselines_corpus_20db}.

\begin{table}[h]
\centering
\caption{In-domain attack WER (\%) for the HuBERT-surrogate-trained generator evaluated on HuBERT-Large-LS960-ft. Clean WER on HuBERT is 2.00--2.06\% across all configurations.}
\label{tab:hubert_indomain}
\begin{tabular}{lccc}
\toprule
Prefix (s) & Delay 0.0\,s & Delay 0.5\,s & Delay 1.0\,s \\
\midrule
1.0 & 21.21 & 27.18 & 19.96 \\
2.0 & 25.17 & 18.63 & 25.42 \\
3.0 & 14.64 & 14.79 & 14.44 \\
4.0 & 14.97 & 21.87 & 12.69 \\
\bottomrule
\end{tabular}
\end{table}

The HuBERT-surrogate generator achieves attack WER between 12.7\% and 27.2\% across all 12 configurations, against a clean baseline of approximately 2.0\%, a 6 to 14 times degradation. The strongest in-domain attack occurs at prefix 1.0\,s, delay 0.5\,s (27.18\% corpus WER), with the strongest length-stratified effect on medium-length utterances of 6--15 words (median $\Delta\mathrm{WER}$ up to 69\%). These numbers are roughly 70--80\% of the corresponding W2V2-surrogate in-domain numbers in Table~\ref{tab:baselines_corpus_20db}, indicating that HuBERT is a moderately harder surrogate to attack but a viable one nonetheless. This validates the HuBERT-surrogate generator as a meaningful basis for the cross-model transferability analysis that follows.

\subsection{Cross-Model Transferability}
\label{app:cross_model_transfer:bidirectional}

We now report cross-model transferability across the full prefix$\times$delay grid in both directions of the W2V2 and HuBERT pair.

\begin{table}[h]
\centering
\caption{Cross-model transfer attack WER (\%) between W2V2 and HuBERT. Each cell shows the attack WER when the generator was trained against one surrogate and the victim is the other model: \textbf{W2V2$\,\to\,$HuBERT} denotes a generator trained against W2V2-Large and evaluated against HuBERT-Large-LS960-ft, and \textbf{HuBERT$\,\to\,$W2V2} denotes a generator trained against HuBERT-Large-LS960-ft and evaluated against W2V2-Large. Clean WER on both victims is approximately 2.0\% across configurations.}
\label{tab:transfer_only}
\begin{tabular}{l ccc ccc}
\toprule
\multirow{2}{*}{\textbf{Prefix (s)}} & \multicolumn{3}{c}{\textbf{W2V2$\,\to\,$HuBERT}} & \multicolumn{3}{c}{\textbf{HuBERT$\,\to\,$W2V2}} \\
\cmidrule(lr){2-4} \cmidrule(lr){5-7}
 & Delay 0.0\,s & Delay 0.5\,s & Delay 1.0\,s & Delay 0.0\,s & Delay 0.5\,s & Delay 1.0\,s \\
\midrule
1.0 & 19.05 & \phantom{0}7.79 & 10.20 & 11.97 & 11.99 & \phantom{0}5.79 \\
2.0 & 19.04 & 10.07 & 10.76 & \phantom{0}3.85 & \phantom{0}9.01 & 11.99 \\
3.0 & 11.57 & 11.52 & \phantom{0}4.98 & \phantom{0}7.41 & \phantom{0}5.80 & \phantom{0}8.06 \\
4.0 & \phantom{0}9.14 & 11.05 & 10.61 & \phantom{0}6.74 & 11.20 & \phantom{0}6.89 \\
\bottomrule
\end{tabular}
\end{table}

\begin{table}[h]
\centering
\caption{Target-window WER (median, \%) under cross-model transfer between W2V2 and HuBERT. Column groups follow the convention defined in Table~\ref{tab:transfer_only}: \textbf{W2V2$\,\to\,$HuBERT} for W2V2-Large-trained generator evaluated against HuBERT-Large-LS960-ft, and \textbf{HuBERT$\,\to\,$W2V2} for the reverse direction. Window WER measures degradation localized to the 3-second perturbed segment.}
\label{tab:transfer_window}
\begin{tabular}{l ccc ccc}
\toprule
\multirow{2}{*}{\textbf{Prefix (s)}} & \multicolumn{3}{c}{\textbf{W2V2$\,\to\,$HuBERT}} & \multicolumn{3}{c}{\textbf{HuBERT$\,\to\,$W2V2}} \\
\cmidrule(lr){2-4} \cmidrule(lr){5-7}
 & Delay 0.0\,s & Delay 0.5\,s & Delay 1.0\,s & Delay 0.0\,s & Delay 0.5\,s & Delay 1.0\,s \\
\midrule
1.0 & 55.6 & 20.0 & 33.3 & 40.0 & 54.6 & 12.5 \\
2.0 & 53.9 & 33.3 & 30.0 & 11.1 & 44.4 & 50.0 \\
3.0 & 30.0 & 30.0 & 12.5 & 33.3 & 25.0 & 28.6 \\
4.0 & 14.3 & 33.3 & 50.0 & 33.3 & 54.6 & 33.3 \\
\bottomrule
\end{tabular}
\end{table}

\paragraph{Bidirectional Transfer.}
The attack transfers effectively in both directions across the W2V2 and HuBERT pair in every configuration evaluated, with corpus-level attack WER between 3.85\% and 19.05\% against a clean baseline of $\sim$2.0\%, a 1.9--9.5$\times$ degradation on a model the generator never trained against. Across all 12 (prefix, delay) configurations, mean transfer attack WER is 11.3\% from W2V2 to HuBERT (range 4.98--19.05\%) and 8.4\% from HuBERT to W2V2 (range 3.85--11.99\%). This bidirectional transfer supports the architectural-family interpretation of CTC-based self-supervised models as a coherent target class: perturbations effective against one model exploit features that generalize to the other.

\paragraph{Where Transfer Is Strongest.} 
For W2V2$\,\to\,$HuBERT, the strongest transfer occurs at short prefixes with zero delay: prefix\,=\,1.0\,s, delay\,=\,0.0\,s achieves 19.05\% corpus WER and 55.6\% median window WER, with 51.0\% of utterances exhibiting $\Delta\mathrm{WER} \geq 0.10$. 
Transfer strength tracks in-domain strength at both ends of the grid: the two strongest in-domain configurations on W2V2 (35.63\% and 32.72\%, Table~\ref{tab:baselines_corpus_20db}) yield the two strongest transfers (19.05\% and 19.04\%), and the two weakest (11.32\% and 16.26\%) yield the two weakest (4.98\% and 7.79\%). In the reverse direction the pattern is less regular, with the highest corpus WERs (11.99\%) occurring at non-zero delays. The length-stratified breakdown reveals that medium-length utterances (6--15 words) are systematically more vulnerable to transferred attacks than long utterances (16+ words), with median $\Delta\mathrm{WER}$ on medium utterances reaching 30--47\% at the strongest transfer configurations in either direction, mirroring the length-sensitivity pattern observed in the in-domain settings.

\paragraph{Directional Asymmetry.} The transfer is asymmetric, with the W2V2 surrogate producing systematically stronger attacks on HuBERT than the reverse. Computing the transfer-retention ratio (transfer attack WER divided by in-domain attack WER) at delay\,=\,0.0\,s, both directions retain roughly comparable proportions of in-domain effectiveness at three of the four prefix lengths (44--58\% retention), but diverge dramatically at prefix\,=\,2.0\,s: W2V2-trained perturbations retain 58\% of their in-domain WER (32.72\% to 19.04\%), while HuBERT-trained perturbations retain only 15\% (25.17\% to 3.85\%). Averaged across all 12 configurations, mean transfer attack WER is 11.3\% from W2V2 to HuBERT versus 8.4\% from HuBERT to W2V2, a 35\% relative gap.

We hypothesize that the asymmetry arises from differences in the two surrogates' pretraining objectives. Both victims are Large models of comparable depth and size, so the asymmetry is unlikely to reflect capacity. W2V2 is trained via a contrastive objective on quantized speech representations, encouraging the encoder to learn linearly separable features at the frame level. HuBERT is trained to predict cluster assignments from offline-discovered acoustic units, which may produce more idiosyncratic decision boundaries that are harder for an external generator to find generalizable perturbations against. Adversarial perturbations crafted to exploit W2V2's more linearly-structured representations may correspondingly transfer more readily to other CTC self-supervised models, while perturbations finding shortcuts in HuBERT's specific cluster geometry are more model-bound. We do not have a definitive mechanistic test of this hypothesis, and we leave a more detailed analysis of surrogate transferability properties to future work.

A practical implication: practitioners deploying CTC-based ASR systems should expect W2V2-derived adversarial inputs to pose a broader threat than equivalent perturbations crafted against HuBERT, even when the deployed model is HuBERT itself.

\paragraph{Universal Baseline as a Reference Point.}
The Universal baseline provides a reference point for how much of this transfer is specific to input-dependent generation. Optimized against W2V2 and evaluated on HuBERT (Table~\ref{tab:universal_to_hubert}), it reaches a mean attack WER of 8.5\% across the 12 configurations, against 11.3\% for SG, with SG ahead in 10 of the 12 cells. The margin has the same shape as in-domain (Table~\ref{tab:baselines_corpus_20db}): it is largest at zero delay (14.7\% against 8.8\%, averaged over prefix lengths) and narrows at non-zero delays. That an input-agnostic perturbation also transfers within the CTC family suggests that the shared vulnerability is a property of the victim family rather than of our generator, a point we return to in Section~\ref{app:cross_model_transfer:whisper}.

\begin{table}[h]
\centering
\caption{Universal baseline under within-family transfer: attack WER (\%) on HuBERT-Large-LS960-ft for the universal perturbation optimized against W2V2-Large, on the same configurations and evaluation subsets as Table~\ref{tab:transfer_only}. Clean WER on HuBERT is approximately 2.0\%. The same perturbation evaluated on Whisper-small is reported in Table~\ref{tab:whisper_transfer_four_methods}.}
\label{tab:universal_to_hubert}
\begin{tabular}{l ccc}
\toprule
\textbf{Prefix (s)} & Delay 0.0\,s & Delay 0.5\,s & Delay 1.0\,s \\
\midrule
1.0 & 10.09 & 10.21 & \phantom{0}9.83 \\
2.0 & \phantom{0}9.61 & \phantom{0}7.07 & \phantom{0}6.79 \\
3.0 & \phantom{0}8.47 & 10.86 & \phantom{0}6.92 \\
4.0 & \phantom{0}7.11 & \phantom{0}7.65 & \phantom{0}7.18 \\
\bottomrule
\end{tabular}
\end{table}

\subsection{Cross-Architecture Transfer to Whisper}
\label{app:cross_model_transfer:whisper}

Within the CTC self-supervised family, our attack achieves substantial transfer (Section~\ref{app:cross_model_transfer:bidirectional}). To probe whether this transferability extends across architectural families, we evaluate against Whisper-small~\citep{radford2022whisper} -- an encoder--decoder seq2seq model with a fundamentally different feature frontend and decoding pathway. We find a sharp architectural cliff: across four black-box attack methods, all converge on attack WER within $\sim$1.5 percentage points of the clean baseline. We emphasize at the outset that this section's purpose is to characterize the boundary at which CTC-trained attacks fail; the regime where our method's advantage over baselines is established 
(in-domain effectiveness on W2V2, Table~\ref{tab:baselines_corpus_20db}, and transfer to HuBERT, where SG exceeds the Universal baseline by about 33\% relative on average; Table~\ref{tab:universal_to_hubert}) is reported in Section~\ref{app:cross_model_transfer:bidirectional} and earlier.

To rule out alternative interpretations of the null transfer to Whisper -- that it reflects a weakness specific to our generator-based attack, the LLM-forecasting step, or the W2V2 surrogate -- we evaluate four black-box attack methods spanning two CTC surrogates and three target-knowledge regimes. Table~\ref{tab:whisper_transfer_four_methods} reports corpus-level attack WER for each method across the full prefix$\times$delay grid.

\begin{table}[h]
\centering
\caption{Cross-architecture transfer attack WER (\%) on Whisper-small for four black-box attack methods. \emph{SG (W2V2)} and \emph{SG (HuBERT)}: our generator with LLM-forecasted target, trained against W2V2 and HuBERT surrogates respectively. \emph{GT}: ground-truth oracle variant of our generator (LLM forecaster replaced with the actual target transcript). \emph{Universal}: input-agnostic universal adversarial perturbation. All methods at SNR=20\,dB. Clean WER on Whisper is 3.91--4.17\% across configurations.}
\label{tab:whisper_transfer_four_methods}
\small
\begin{tabular}{cc cccc}
\toprule
\textbf{Prefix (s)} & \textbf{Delay (s)} & \textbf{SG (W2V2)} & \textbf{SG (HuBERT)} & \textbf{GT (W2V2)} & \textbf{Universal\ (W2V2)} \\
\midrule
\multirow{3}{*}{1.0} & 0.0 & 4.90 & 4.76 & 5.31 & 4.87 \\
                     & 0.5 & 4.48 & 5.15 & 4.55 & 5.00 \\
                     & 1.0 & 4.40 & 4.21 & 4.56 & 4.52 \\
\addlinespace
\multirow{3}{*}{2.0} & 0.0 & 4.50 & 4.27 & 4.57 & 4.78 \\
                     & 0.5 & 4.38 & 4.48 & 4.43 & 4.74 \\
                     & 1.0 & 4.43 & 4.36 & 4.52 & 4.48 \\
\addlinespace
\multirow{3}{*}{3.0} & 0.0 & 4.31 & 4.48 & 4.24 & 4.55 \\
                     & 0.5 & 4.33 & 4.36 & 4.52 & 4.57 \\
                     & 1.0 & 4.16 & 4.18 & 4.26 & 4.42 \\
\addlinespace
\multirow{3}{*}{4.0} & 0.0 & 4.10 & 4.35 & 4.31 & 4.32 \\
                     & 0.5 & 4.19 & 4.11 & 4.17 & 4.44 \\
                     & 1.0 & 4.23 & 4.37 & 4.21 & 4.30 \\
\midrule
\multicolumn{2}{c}{\textbf{Mean}} & \textbf{4.37} & \textbf{4.39} & \textbf{4.47} & \textbf{4.58} \\
\bottomrule
\end{tabular}
\end{table}

Across all 12 configurations and all four attack methods, attack corpus WER on Whisper falls within $\sim$1.5 percentage points of the clean baseline of approximately 4\%, with mean $\Delta\mathrm{WER}$ ranging from 0.33\% (SG with W2V2 surrogate) to 0.54\% (Universal).\footnote{Whisper's clean WER on the LibriSpeech \texttt{test-clean} subset is higher than W2V2's or HuBERT's primarily because of occasional hallucinated tokens on short audio segments rather than systematic transcription errors.} Window-level metrics are similarly unmoved: median target-window WER is 0\% in all 48 (cell, method) combinations, and ASR@$0.10$ remains below 9\% across the entire grid.

The convergent null result across four attack variants points to the victim architecture, rather than the attack design, as the dominant factor. Three orthogonal manipulations of the attack pipeline leave the null transfer intact: (i) substituting the LLM forecaster with oracle access to the target transcript (GT) does not improve transfer, ruling out the LLM-forecasting step as the bottleneck; 
(ii) substituting our generator-based attack with a generic input-agnostic universal perturbation, which does transfer within the CTC family (8.5\% on HuBERT; Table~\ref{tab:universal_to_hubert}), does not improve transfer to Whisper, ruling out a method-specific overfitting to W2V2; (iii) substituting the W2V2 surrogate with HuBERT does not improve transfer, ruling out a surrogate-specific shortcut. The CTC to seq2seq robustness gap is therefore a property of the victim architecture family rather than of any particular attack design or surrogate choice.

Three architectural factors plausibly explain the absence of transfer to Whisper. First, Whisper is an encoder--decoder model with autoregressive token generation conditioned on a strong implicit language model, which can override locally corrupted acoustic evidence using surrounding context. Second, Whisper's log-mel feature frontend and BPE tokenizer differ substantially from the raw-waveform CNN and character-level CTC heads of W2V2 and HuBERT, so perturbations crafted to manipulate frame-level CTC logits have no direct analogue in Whisper's representational space. Third, Whisper's encoder attends bidirectionally over a 30\,s window, so the clean audio on both sides of the 3.0\,s target window remains available when decoding the corrupted segment.

Combined with the strong bidirectional within-CTC transfer reported in Section~\ref{app:cross_model_transfer:bidirectional}, the convergent results in Table~\ref{tab:whisper_transfer_four_methods} support an interpretation of attacks crafted against CTC self-supervised representations as exploiting acoustic features specific to that architectural family rather than generic distortions affecting any ASR system. We do not claim our attack is uniquely strong against seq2seq victims; the more substantive open question is whether \emph{any} generator-based attack can bridge the CTC to seq2seq boundary, for instance via ensemble-surrogate training that incorporates a seq2seq model in the training loop. This and the related question of white-box attack ceilings against Whisper are natural extensions we leave to future work.

\subsection{Implications}
\label{app:cross_model_transfer:implications}

The transferability profile suggests several conclusions for both attack design and threat-model analysis. First, an attacker need not have exact knowledge of the deployed CTC ASR model: a single CTC surrogate suffices to threaten an entire family of self-supervised CTC systems, with attack effectiveness reduced but not eliminated under transfer in either direction. Second, the choice of surrogate matters substantively even within the CTC family: W2V2 produces more transferable perturbations than HuBERT despite both belonging to the same architectural family. Third, defenders deploying seq2seq ASR systems such as Whisper appear to benefit from substantial robustness to perturbations crafted against CTC surrogates. Whether this advantage extends to perturbations crafted via \emph{ensemble surrogate training}, in which the generator would be jointly optimized against W2V2, HuBERT, and Whisper losses, is a natural extension we leave to future work.

\end{document}